\documentclass[letterpaper]{article} % DO NOT CHANGE THIS
\usepackage{aaai25}  % DO NOT CHANGE THIS
\usepackage{times}  % DO NOT CHANGE THIS
\usepackage{helvet}  % DO NOT CHANGE THIS
\usepackage{courier}  % DO NOT CHANGE THIS
\usepackage[hyphens]{url}  % DO NOT CHANGE THIS
\usepackage{graphicx} % DO NOT CHANGE THIS
\usepackage{subfigure}
\usepackage{natbib}  % DO NOT CHANGE THIS AND DO NOT ADD ANY OPTIONS TO IT
\usepackage{caption} % DO NOT CHANGE THIS AND DO NOT ADD ANY OPTIONS TO IT
\usepackage{algorithm}
\usepackage{algorithmic}
\usepackage{amsmath}
\usepackage{amssymb}
\usepackage{mathtools}
\usepackage{amsthm}
\usepackage{tablefootnote}
\usepackage{diagbox}

\usepackage{newfloat}
\usepackage{listings}
\DeclareCaptionStyle{ruled}{labelfont=normalfont,labelsep=colon,strut=off} % DO NOT CHANGE THIS
\floatstyle{ruled}
\newfloat{listing}{tb}{lst}{}
\floatname{listing}{Listing}
\theoremstyle{plain}
\newtheorem{theorem}{Theorem}
\newtheorem{proposition}{Proposition}

\newtheorem{assumption}{Assumption}
\newtheorem{definition}{Definition}
\theoremstyle{remark}

\newcommand{\X}{\rmX}
\newcommand{\Y}{\rmY}
\newcommand{\F}{\gF}
\newcommand{\I}{\sI}
\newcommand{\p}{\sP}

\def\Figref#1{Figure~\ref{#1}}

\def\eqref#1{equation~(\ref{#1})}
\def\Eqref#1{Equation~(\ref{#1})}
\def\Algref#1{Algorithm~\ref{#1}}

\def\1{\bm{1}}

\def\eps{{\epsilon}}

\def\rs{{\textnormal{s}}}

\def\rvp{{\mathbf{p}}}

\def\rmA{{\mathbf{A}}}

\def\rmX{{\mathbf{X}}}
\def\rmY{{\mathbf{Y}}}

\DeclareMathAlphabet{\mathsfit}{\encodingdefault}{\sfdefault}{m}{sl}
\SetMathAlphabet{\mathsfit}{bold}{\encodingdefault}{\sfdefault}{bx}{n}

\def\gA{{\mathcal{A}}}

\def\gF{{\mathcal{F}}}

\def\gH{{\mathcal{H}}}

\def\gS{{\mathcal{S}}}

\def\gU{{\mathcal{U}}}

\def\sF{{\mathbb{F}}}

\def\sI{{\mathbb{I}}}

\def\sP{{\mathbb{P}}}

\def\sR{{\mathbb{R}}}

\def\sT{{\mathbb{T}}}

\newcommand{\E}{\mathbb{E}}

\newcommand{\R}{\mathbb{R}}

\newcommand{\KL}{D_{\mathrm{KL}}}

\title{Scalable Acceleration for Classification-Based Derivative-Free Optimization}
\author{
    Tianyi Han\textsuperscript{\rm *},
    Jingya Li,
    Zhipeng Guo,
    Yuan Jin\textsuperscript{\rm *} 
}
\affiliations{
    \{ty.han\}\{jy.li\}\{zp.guo\}\{y.jin\}@supreium.com \\
    Beijing Supreium Technology \\
    No.35 Tsinghua East Road \\ 
    Haidian District, Beijing, China
}

\usepackage{bibentry}
\begin{document}

\maketitle

\begin{abstract}
    Derivative-free optimization algorithms play an important role in scientific and engineering design optimization problems, especially when derivative information is not accessible. In this paper, we study the framework of sequential classification-based derivative-free optimization algorithms. By introducing learning theoretic concept hypothesis-target shattering rate, we revisit the computational complexity upper bound of \emph{SRACOS} \cite{hu2017sequential}. Inspired by the revisited upper bound, we propose an algorithm named \emph{RACE-CARS}, which adds a random region-shrinking step compared with \emph{SRACOS}. We further establish theorems showing the acceleration by region shrinking. Experiments on the synthetic functions as well as black-box tuning for language-model-as-a-service demonstrate empirically the efficiency of \emph{RACE-CARS}. An ablation experiment on the introduced hyper-parameters is also conducted, revealing the mechanism of \emph{RACE-CARS} and putting forward an empirical hyper-parameter tuning guidance.
\end{abstract}


\section{Introduction}
\label{introduction}

In recent years, there has been a growing interest in the field of derivative-free optimization (DFO) algorithms, also known as zeroth-order optimization. These algorithms aim to optimize objective functions without relying on explicit gradient information, making them suitable for scenarios where obtaining derivatives is either infeasible or computationally expensive \cite{conn2009introduction,larson2019derivative}. For example, DFO techniques can be applied to hyperparameter tuning, which involves optimizing complex objective functions with unavailable first-order information \cite{falkner2018bohb,akiba2019optuna,yang2020hyperparameter}. Moreover, DFO algorithms find applications in engineering design optimization, where the objective functions are computationally expensive to evaluate and lack explicit derivatives \cite{ray2001engineering,liao2010two,akay2012artificial}.
 
Classical DFO methods such as Nelder-Mead method \cite{nelder1965simplex} or directional direct-search (DDS) method \cite{cea1971optimisation,yu1979positive} have shown great success on convex problems. However, their performance is compromised when the objective is nonconvex, which is the commonly-faced situation for black-box problems. In this paper, we focus on optimization problems reading as
\begin{equation}\label{objmodel}
    \min_{x\in\Omega}~f(x),
\end{equation}
where the $n$ dimensional compact cubic $\Omega\subseteq\sR^n$ is solution space. In addition, we will not stipulate any convexity, smoothness or separability assumption on $f.$ 

Recent decades, progresses have been made in the extensive exploration of DFO algorithms for nonconvex problems, and various kinds of algorithms have been established. For example, evolutionary algorithms \cite{back1993overview,fortin2012deap,hansen2016cma,opara2019differential}, Bayesian optimization (BO) methods \cite{snoek2012practical,shahriari2015taking,frazier2018tutorial} and gradient approximation surrogate modeling algorithms \cite{nesterov2017random,chen2019zo,ge2022solnp+,ragonneau2023pdfo}.

Given the \emph{nonconvex} and \emph{black-box} nature of the problem, the quest for enhanced sample efficiency inevitably presents the classic trade-off between exploration and exploitation. Similar to typical nonlinear/nonconvex numerical issues, the aforementioned algorithms are also susceptible to the curse of dimensionality \cite{bickel2004some,hall2005geometric,fan2008high,shi2021numerical,scheinberg2022finite,yue2023zeroth}. Improving scalability and efficiency can be distilled into a few key areas, such as function decomposition \cite{wang2018batched}, dimension reduction \cite{wang2016bayesian,nayebi2019framework} and sample strategy refinement. Regarding the latter, previous algorithms have focused on preventing over-exploitation through, for instance, trying more efficient sampling dynamics \cite{ros2008simple,hensman2013gaussian,yi2024improving}, search space discretization \cite{sobol1967distribution} or search region restriction \cite{eriksson2019scalable,wang2020learning}.

Regardless of the improvements, many of these model-based DFO algorithms share the mechanism that, locally or globally, fits the objective function. The motivation of classification-based DFO algorithms is different: it learns a hypothesis (classifier) to fit the sub-level set $\Omega_{\alpha_t}:=\{x\in\Omega\colon f(x)-f^*\le\alpha_t\}$ at each iteration $t=1,\ldots,T,$ with which new candidates are generated for the successive round \cite{michalski2000learnable,gotovos2013active,bogunovic2016truncated,yu2016derivative,hashimoto2018derivative}. Sub-level sets are considered to contain less information than objective functions, since they are already obtained whenever the objective function is known. By this means, sample usage is more efficient. Moreover, it is less sensitive to function oscillation or noise \cite{liu2017zoopt} and easy to be parallelized \cite{liu2017zoopt,liu2019asynchronous,hashimoto2018derivative}. 

Training strategies of hypotheses have branched out in various directions. \citeauthor{hashimoto2018derivative} pursue accurate hypotheses using a \emph{conservative} strategy raised by \citeauthor{el2012active}. They demonstrate that for a hypothesis class $\mathcal{H}$ with a VC-dimension of $VC(\mathcal{H}),$ an $\epsilon$-accurate hypothesis requires $\mathcal{O}(\epsilon^{-1}(VC(\mathcal{H})\log(\epsilon^{-1})))$ samples per batch. While a computationally efficient approximation has indeed been developed, its success was demonstrated only in the context of low-dimensional problems. The extent of its effectiveness in high-dimensional scenarios is yet to be determined. However, a contrasting approach successfully lead to the improvements in sample efficiency. \citeauthor{yu2016derivative} propose to ``RAndomly COordinate-wise Shrink'' the solution region (\emph{RACOS}), a more \emph{radical} approach that diverges from the goal of producing accurate hypotheses. They prove \emph{RACOS} converges to global minimum in polynomial time when the objective is locally Holder continuous, and experiments indicate success on high-dimensional problems ranging from hundreds to thousands of dimensions. An added benefit is the ease of sampling through the generated hypotheses, as their active regions are coordinate-orthogonal cubics. Building on this, the sequential counterpart of \emph{RACOS}, known as \emph{SRACOS}, takes an even more radical stance by sampling only once per iteration and learning from biased data. Despite this, \emph{SRACOS} has been shown to outperform \emph{RACOS} under certain mild restriction \cite{hu2017sequential}.

\subsection{Outline and Contributions}

\begin{itemize}
    \item [(1)] Our research extends the groundwork laid by \cite{yu2016derivative,hu2017sequential}, yet we discover that the upper bound they proposed for the \emph{SRACOS} does not fully encapsulate its behavior, potentially allowing for an overflow. On the other hand, we construct a counter-example (eq. \ref{counter}) illustrating the upper bound is not tight enough to delineate the convergence accurately.
    \item [(2)] We also identify a contentious assumption regarding the concept of error-target dependence that seems to underpin these limitations. In this paper, we introduce a novel learning theoretic concept termed the hypothesis-target $\eta$-shattering rate, which serves as a foundation for our reassessment of the upper bound on the computational complexity of \emph{SRACOS}.
    \item [(3)]Inspired by the reanalysis, we recognize that the accuracy of the trained hypotheses is even less critical than previously thought. With this insight, we formulate a novel algorithm, \emph{RACE-CARS} (an acronym for ``RAndomized CoordinatE Classifying And Region Shrinking"), which perpetuates the essence of \emph{SRACOS} while promises a theoretical enhancement in convergence. We design experiments on both synthetic functions and black-box tuning for LMaaS. These experiments juxtapose \emph{RACE-CARS} against a selection of DFO algorithms, empirically highlighting the superior efficacy of \emph{RACE-CARS}.
    \item [(4)] In discussion and appendix, we further substantiate the empirical prowess of \emph{RACE-CARS} beyond the confines of continuity, also supported by theoretical framework. An ablation study is also conducted to demystify the selection of hyperparameters for our proposed algorithm.
\end{itemize}

The rest of the paper is organized in five sections, sequentially presenting the background, theoretical study, experiments, discussion and conclusion.
 
\section{Background}

\begin{assumption}
    In eq. \ref{objmodel}, we presume $f(x)$ is lower bounded within $\Omega$, with $f^*:=\min_{x\in\Omega}~f(x).$
\end{assumption}
In our theoretical analysis, we refine the procedure as a stochastic process. Let $\gF$ denote the Borel $\sigma$-algebra on $\Omega$, and $\sP$ the probability measure defined on $\gF.$ For instance, if $\Omega$ is continuous space, $\sP$ is derived from Lebesgue measure $m$, such that $\sP(B):=m(B)\slash m(\Omega)$ for all $B\in\gF.$
\begin{assumption}
    Define $\Omega_\epsilon:=\{x\in\Omega\colon f(x)-f^*\le\epsilon\}$, with the assumption $|\Omega_\epsilon|:=\sP(\Omega_\epsilon)>0$ for all $\epsilon>0.$
\end{assumption}

A hypothesis (or classifier) $h$ is a function mapping the solution space $\Omega$ to $\{0, 1\}.$ Define 
\begin{equation}\label{stochasticprocess}
   D_h(x):=
   \begin{cases}
      \sP(\{x\in\Omega\colon h(x)=1\})^{-1},&h(x)=1\\
      0,&\text{otherwise},
   \end{cases}
\end{equation}
a probability distribution in $\Omega.$ Let $\rmX_h$ be the random vector in $\left(\Omega,\gF,\sP\right)$ drawn from $D_h,$ implying that $\Pr(\rmX_h\in B)=\int_{x\in B} D_h(x)d\sP$ for all $B\in\gF.$ Denoted by $\sT:=\{1,2,\ldots,T\}$, and let $\mathbb{F}:=\left(\gF_t\right)_{t\in\sT}$ be a filtration of $\sigma$-algebras on $\Omega$ indexed by $\sT$, satisfying $\gF_1\subseteq\gF_2\subseteq\cdots\subseteq\gF_T\subseteq\gF.$ A typical classification-based optimization algorithm learns an $\sF$-adapted stochastic process $\rmX:=(\rmX_t)_{t\in\sT},$ where each $\rmX_t$ is induced by $\rmX_{h_t}$ and $h_t$ is the hypothesis updated at step $t.$ Subsequently, it samples new data with a stochastic process $\rmY:=(\rmY_t)_{t\in\sT}$ generated by $\rmX$. Typically, the new candidates at step $t\ge r+1$ are sampled from
\begin{equation*}
   \rmY_t:=
    \begin{cases}
      \rmX_t, &\text{with probability $\lambda$}\\
      \rmX_\Omega, &\text{with probability $1-\lambda$},
    \end{cases}
\end{equation*}
where $\rmX_\Omega$ is random vector drawn from uniform distribution $\gU_\Omega$ and $\lambda \in [0, 1]$ is the exploitation rate. 

A simplified batch-mode classification-based optimization algorithm is outlined in \Algref{classification}. At each step $t,$ it selects a positive set $\gS_{positive}$ from $\gS$ containing the best $m$ samples,  with the remainder in the negative set $\gS_{negative}$. Then it trains a hypothesis $h_t$ distinguishing between the positive set and negative set, ensuring $h_t(x_j)=0$ for all $(x_j, y_j)\in\gS_{negative}.$ Finally, it samples $r$ new solutions with the sampling random vector $\rmY_t.$ The sub-procedure $h_t\leftarrow\mathcal{T}(\mathcal{S}_{positive},\mathcal{S}_{negative})$ means hypothesis training.

\begin{algorithm}[htb]
    \caption{Batch-Mode Classification-Based Optimization Algorithm}
    \label{classification}
    \begin{algorithmic}
        \STATE {\bfseries Input:} $T$: Budget; $r$: Training size; $m$: Positive size.
        \STATE {\bfseries Output: }$(x_{best},y_{best}).$
        \STATE Collect: $\gS=\{(x^0_1,y^0_1),\ldots,(x^0_r,y^0_r)\}$ i.i.d. from $\gU_\Omega$;
        \STATE $(x_{best},y_{best})=\arg\min\{y\colon(x,y)\in\gS\}$;
        \FOR{$t=1,\ldots,T\slash r$}
            \STATE Classify: $(\mathcal{S}_{positive},\mathcal{S}_{negative})\leftarrow\mathcal{S}$;
            \STATE Train: $h_t\leftarrow\mathcal{T}(\mathcal{S}_{positive},\mathcal{S}_{negative})$;
            \STATE Sample: $\{(x^t_1,y^t_1),\ldots,(x^t_r,y^t_r)\}$ i.i.d. with $\Y_t$;
            \STATE Select: $\mathcal{S}\leftarrow \mathcal{S}\cup \{(x^t_1,y^t_1),\ldots,(x^t_r,y^t_r)\}$;
            \STATE $(x_{best},y_{best})=\arg\min\bigl\{y\colon(x,y)\in\mathcal{S}\bigr\}.$
        \ENDFOR
        \STATE {\bfseries Return:} $(x_{best},y_{best})$
    \end{algorithmic}
\end{algorithm}

\emph{RACOS} is the abbreviation of ``RAndomized COordinate Shrinking''. Literally, it trains the hypothesis by this means \cite{yu2016derivative}, i.e., shrinking coordinates randomly such that all negative samples are excluded from active region of resulting hypothesis. \Algref{racos} shows a continuous version of \emph{RACOS}.

\begin{algorithm}[htb]
    \caption{RACOS}
    \label{racos}
    \begin{algorithmic}
        \STATE {\bfseries Input:} $\Omega$: Boundary; $(\mathcal{S}_{positive},\mathcal{S}_{negative})$: Binary sets; $\sI=\{1,\ldots,n\}$: Index of dimensions.
        \STATE {\bfseries Output:} $h$: Hypothesis.
        \STATE Randomly select: $x_+=(x^1_+,\ldots,x^n_+)\leftarrow\gS_{positive}$;
        \STATE Initialize: $h(x)\equiv 1$;
        \WHILE{$\exists x\in\mathcal{S}_{negative}$ s.t. $h(x)=1$}
            \STATE Randomly select: $k\leftarrow \sI$;
            \STATE Randomly select: $x_-=(x^1_-,\ldots,x^n_-)\leftarrow \mathcal{S}_{negative}$;
            \IF{$x^k_+\le x^k_-$}
                \STATE $\rs\leftarrow random(x^k_+,x^k_-)$;
                \STATE Shrink: $h(x)=0,~\forall x\in\{x=(x^1,\ldots,x^n)\in\Omega\colon x^k>\rs\}$;
            \ELSE
                \STATE $\rs\leftarrow random(x^k_-,x^k_+)$;
                \STATE Shrink: $h(x)=0,~\forall x\in\{x=(x^1,\ldots,x^n)\in\Omega\colon x^k<\rs\}$;
            \ENDIF
        \STATE {\bfseries Return:} $h$
        \ENDWHILE
    \end{algorithmic}
\end{algorithm}
Aside from sampling only once per iteration, another difference between sequential and batch mode classification-based DFO is that the sequential version will replace the training set with a new one under certain rules to finish step $t$ (see in in \Algref{sequentialclassification}). In the rest of this paper, we will omit the details of $Replacing$ sub-procedure which can be found in \cite{hu2017sequential}.

\begin{algorithm}[htb]
    \caption{Sequential-Mode Classification-Based Optimization Algorithm}
    \label{sequentialclassification}
    \begin{algorithmic}
        \STATE {\bfseries Input:} $T$: Budget; $r$: Training size; $m$: Positive size; $Replacing$: Replacing sub-procedure.
        \STATE {\bfseries Output:} $(x_{best},y_{best}).$
        \STATE Collect $\mathcal{S}=\{(x_1,y_1),\ldots,(x_r,y_r)\}$ i.i.d. from $\gU_\Omega$;
        \STATE $(x_{best},y_{best})=\arg\min\{y\colon(x,y)\in\mathcal{S}\}$;
        \FOR{$t=r+1,\ldots,T$}
            \STATE Classify: $(\mathcal{S}_{positive},\mathcal{S}_{negative})\leftarrow\mathcal{S}$;
            \STATE Train: $h_t\leftarrow\mathcal{T}(\mathcal{S}_{positive},\mathcal{S}_{negative})$;
            \STATE Sample: $(x_t,y_t)\sim \rmY_t$;
            \STATE Replace: $\mathcal{S}\leftarrow Replacing((x_t,y_t),\mathcal{S})$;
            \STATE $(x_{best},y_{best})=\arg\min\{y\colon(x,y)\in\mathcal{S}\}$;
        \ENDFOR
        \STATE {\bfseries Return:} $(x_{best},y_{best})$
    \end{algorithmic}
\end{algorithm}
 
Classification-based DFO algorithms admit a bound on the query complexity \cite{yu2014sampling}, quantifying the total number of function evaluations required to identify a solution that achieves an approximation level of $\eps$ with a high probability of at least $1-\delta.$
\begin{definition}[$(\epsilon,\delta)$-Query Complexity]{\rm\label{convergecencept}
   Given $f,$ $0<\delta<1$ and $\epsilon>0.$ The $(\epsilon,\delta)$-query complexity of an algorithm $\gA$ is the number of calls to $f$ such that, with probability at least $1-\delta,$ $\gA$ finds at least one solution $\tilde{x}\in\Omega$ satisfying
   \begin{equation*}
       f(\tilde{x})-f^*\le\epsilon.
   \end{equation*}
}
\end{definition}

Definition \ref{errortarget} and \ref{gammashrinking} are given by \cite{yu2016derivative}. The first one characterizes the so-called dependence between classification error and target region, which is expected to be minimized to ensure that the efficiency. The second one characterizes the portion of active region of hypothesis, and is also expected to be as small as possible.

\begin{definition}[Error-Target $\theta$-Dependence]{\rm \label{errortarget}
    The error-target dependence $\theta\ge0$ of a classification-based optimization algorithm is its infimum such that, for any $\epsilon>0$ and any $t=1,\ldots,T,$
    \begin{equation*}
        \big{|}|\Omega_\epsilon|\cdot\sP(\mathcal{R}_t)-\sP\bigl(\Omega_\epsilon\cap\mathcal{R}_t\bigr)\big{|}\le\theta|\Omega_\epsilon|,
    \end{equation*}
    where $\mathcal{R}_t:=\Omega_{\alpha_t}\Delta\{x\in\Omega\colon h_t(x)=1\}$ denotes the relative error, $\Omega_{\alpha_t}$ is the sub-level set at step $t$ with $\alpha_t:=\min\limits_{1\le i\le t}f(x_i)-f^*$ and the operator $\Delta$ is symmetric difference of two sets defined as $A_1\Delta A_2=(A_1\cup A_2)-(A_1\cap A_2).$ 
}    
\end{definition}

\begin{definition}[$\gamma$-Shrinking Rate]\label{gammashrinking}{\rm
    The shrinking rate $\gamma>0$ of a classification-based optimization algorithm is its infimum such that $\sP(x\in\Omega\colon h_t(x)=1)\le\gamma|\Omega_{\alpha_t}|,$ for all $t=1,\ldots,T.$
}
\end{definition}
 
\section{Theoretical Study}\label{theostudy}

Previous studies gave a general bound of the query complexity of \Algref{sequentialclassification} based on minor error-target $\theta$-dependence and $\gamma$-shrinking rate assumptions:
\begin{theorem}\label{srcoscomplexity}{\rm\cite{hu2017sequential}
    Given $0<\delta<1$ and $\epsilon>0,$ if a sequential classification-based optimization algorithm has error-target $\theta$-dependence and $\gamma$-shrinking rate, then its $(\epsilon,\delta)$-query complexity is upper bounded by
    \begin{equation*}
        \mathcal{O}\biggl(\max\left\{\frac{1}{|\Omega_\epsilon|}\bigl(\lambda+\frac{1-\lambda}{\gamma(T-r)}\sum^{T}_{t=r+1}\Phi_t\bigr)^{-1}\ln\frac{1}{\delta},T\right\}\biggr),
    \end{equation*}
    where
    $\Phi_t=\biggl(1-\theta-\sP(\mathcal{R}_{D_t})-m(\Omega)\sqrt{\frac{1}{2}\KL(D_t\|\gU_\Omega)}\biggr)\cdot|\Omega_{\alpha_t}|^{-1}$
    with the notations $D_t:=\lambda D_{h_t}+(1-\lambda)\gU_\Omega$ and $\sP(\mathcal{R}_{D_t}):=\int_{\mathcal{R}_t}D_td\sP.$
}
\end{theorem}

\subsection{Issues Introduced by Error-Target Dependence}\label{deficiencies}

\begin{itemize}
    \item [{\bf (1)}] {\bf Overflow of the upper bound}
        
    As assumptions entailed in \cite{yu2016derivative,hu2017sequential}, it can be observed that lower values of $\theta$ or $\gamma$ correlate with improved query complexity. However, this is not a hard-and-fast rule. Even with small values of these parameters, we can encounter scenarios where the expected performance does not materialize. Following the lemma given by \cite{yu2016derivative}: $\sP(\mathcal{R}_t)\le\sP(\mathcal{R}_{D_t})+m(\Omega)\sqrt{\frac{1}{2}\KL(D_t\|\gU_\Omega)}$, we have the following inequality:
    \begin{align*}
    \Phi_t=&\biggl(1-\theta-\sP(\mathcal{R}_{D_t})-m(\Omega)\sqrt{\frac{1}{2}\KL(D_t\|\gU_\Omega)}\biggr)\cdot|\Omega_{\alpha_t}|^{-1}\\
    \le&(1-\theta-\sP(\mathcal{R}_t))\cdot|\Omega_{\alpha_t}|^{-1}.
    \end{align*}
    The concept of error-target $\theta$-dependence reveals that a small $\theta$ does not guarantee small relative error $\sP(\mathcal{R}_t).$ Contrarily, a small $\theta$ coupled with a large $(\sP(\Omega_\epsilon\cap\mathcal{R}_t))\slash|\Omega_\epsilon|$ can result in a significant $\sP(\mathcal{R}_t),$ which can even be $1$ as long as $\Omega_{\alpha_t}$ is totally out of the active region of $h_t,$ when the hypothesis $h_t$ is completely wrong. Problematically, as a divisor in the proof of Theorem \ref{srcoscomplexity}, $\Phi_t\le(1-\theta-\sP(\mathcal{R}_t))\cdot|\Omega_{\alpha_t}|^{-1}$ is less or equal to 0, disrupting the established inequality. Nevertheless, in order that the inequality disrupt, $\sP(\mathcal{R}_t)$ is not necessary to be $1$ as $\theta$ is inherently nonnegative, highlighting that a series of inaccurate hypotheses can undermine the validity of the upper bound. This finding challenges the principle of \emph{SRACOS}'s radical training strategy.

    \item [{\bf (2)}] {\bf Tightness of the upper bound}
        
    Consider an extreme but plausible situation where the hypotheses generated at each step are defined as follows:
    \begin{equation}\label{counter}
        h_t(x)=
        \begin{cases}
            1, &x\in\Omega_\epsilon\\
            0, &x\notin\Omega_\epsilon.
        \end{cases}
    \end{equation}
    In the context of sequential-mode classification-based optimization algorithms, where the training sets are not only small but also potentially biased, it's reasonable to expect large relative errors $\sP(\mathcal{R}_t)$. This scenario could lead to a series of hypotheses that are inaccurate with respect to $\Omega_{\alpha_t}$ but, by chance, accurate with respect to $\Omega_\epsilon.$ Consequently, the error-target dependence $\theta=\max_{1\le t\le T}\sP(\mathcal{R}_t)$ can be unexpectedly large. Even all $\Phi_t$ values being positive, the query complexity bound given in Theorem \ref{srcoscomplexity} is therefore not optimistic. However, the probability of failing to identify an $\epsilon$-minimum:
    \begin{align*}
        &\Pr\bigl(\min_{1\le t\le T}f(x_t)-f^*\ge\epsilon\bigr)\\
        =&(1-|\Omega_\epsilon|)^r\biggl((1-\lambda)(1-|\Omega_\epsilon|)\biggr)^{T-r},
    \end{align*}
    is less than $\delta$ for a reasonably sized $T$. This indicates that the upper bound may not be as tight as initially thought.
\end{itemize}

\subsection{Revisit of Query Complexity Upper Bound}

It's evident that even minimal error-target dependence can not encapsulate issues arising from substantial relative error. This is because error-target dependence alone is insufficient to fully account for relative error. Intuitively, one might consider introducing an additional assumption to cap relative error. However, such an assumption would be impractical, given the inherently small and biased nature of training datasets in the process. To this end, we give a new concept that stands apart from the constraints of relative error:

\begin{definition}[Hypothesis-Target $\eta$-Shattering Rate]\label{hypotarget}{\rm
    Given $\eta\in[0,1],$ for a family of hypotheses $\gH$ defined on $\Omega,$ we say $\Omega_\epsilon$ is $\eta$-shattered by $h\in\gH$ if

    \begin{equation*}
        \sP(\Omega_\epsilon\cap \{x\in\Omega\colon h(x)=1\})\ge\eta|\Omega_\epsilon|,
    \end{equation*}
    and $\eta$ is called hypothesis-target shattering rate.
}
\end{definition}

The hypothesis-target shattering rate mirrors the error-target dependence in its relation to the hypothesis's target-accuracy. Importantly, it provides a limitation for error-target dependence in conjunction with relative error:
$$\theta\le\max\{\sP\bigl(\mathcal{R}_t\bigr),|1-\sP\bigl(\mathcal{R}_t\bigr)-\eta|\}.$$ This rate, $\eta$, measures the overlap between the target set $\Omega_\epsilon$ and the active region of a hypothesis. Crucially, it also mitigates the influence of relative error on error-target dependence. Utilizing the concept hypothesis-target shattering rate, we reexamine the upper bound of $(\epsilon,\delta)$-query complexity in the subsequent theorem.

\begin{theorem}\label{sracosconvergence}{\rm
    For sequential-mode classification-based DFO \Algref{sequentialclassification}, let $\rmX_t=\rmX_{h_t},$ $\epsilon>0$ and $0<\delta<1.$ When $\Omega_\epsilon$ is $\eta$-shattered by $h_t$ for all $t=r+1\ldots,T$ and $\max\limits_{t=r+1,\ldots,T}\sP(\{x\in\Omega\colon h_t(x)=1\})\le p\le 1,$ the $(\epsilon,\delta)$-query complexity is upper bounded by 
    \begin{equation*}
        \mathcal{O}\bigg(\max\{\bigl(\lambda\frac{\eta}{p}+(1-\lambda)\bigr)^{-1}\bigl(\frac{1}{|\Omega_\epsilon|}\ln\frac{1}{\delta}-r\bigr)+r,T\}\bigg).
    \end{equation*}
}
\end{theorem}

\subsection{The Region-Shrinking Acceleration}\label{acceleration}

In the analysis of $(\epsilon, \delta)$-query complexity for classification-based optimization, the focus shifts away from minimizing relative error $\sP(\mathcal{R}_t)$, as our goal is to identify optima, not to develop a sequence of accurate hypotheses. The counter-example in \eqref{counter}, despite a potentially high relative error, represents an optimal hypothesis scenario. This realization allows for the consideration of more radical hypotheses, directing our attention to the overlap between $\Omega\epsilon$ and the active region of the hypotheses, which is quantified by the hypothesis-target shattering rate.

The $\gamma$-shrinking rate, as defined in Definition \ref{gammashrinking}, measures the decay of $\sP(x\in\Omega \colon h_t(x)=1)$. However, the rapid decrease of $|\Omega_{\alpha_t}|$ as $\alpha_t$ approaches zero makes it impractical to sustain a series of hypotheses with a small $\gamma$ through our training process. Thus, the $\gamma$-shrinking assumption is often not feasible for minimal $\gamma$.

Moving beyond the pursuit of minimal relative error and $\gamma$-shrinking relative to $|\Omega_{\alpha_t}|$, we introduce Algorithm \ref{racecars}, which adaptively shrinks the sampling random vector $\rmY_t$'s active region through a $Projection$ sub-procedure.

\begin{algorithm}[htb]
    \caption{Accelerated Sequential-Mode Classification-Based Optimization Algorithm}
    \label{racecars}
    \begin{algorithmic}
        \STATE {\bfseries Input:} $\Omega$: Boundary; $T\in\mathbb{N}^+$: Budget; $\quad r=m+k$; $Replacing$: Replacing sub-procedure;
        $\gamma$: Region shrinking rate; $\rho$: Region shrinking frequency.
        \STATE {\bfseries Output:} $(x_{best},y_{best}).$
        \STATE Collect $\mathcal{S}=\{(x_1,y_1),\ldots,(x_r,y_r)\}$ i.i.d. from $\gU_\Omega$;
        \STATE $(x_{best},y_{best})=\arg\min\{y\colon(x,y)\in\mathcal{S}\}$;
        \STATE Initialize $k=1,$ $\tilde{\Omega}=\Omega$;
        \FOR{$t=r+1,\ldots,T$}
            \STATE Train: $h_t\leftarrow\mathcal{T}(\mathcal{S}_{positive},\mathcal{S}_{negative})$;
            \STATE $\rs\leftarrow random(0,1)$;
            \IF{$\rs\le\rho$}
                \STATE Shrink region: $\tilde{\Omega}=\Omega\cap[x_{best}-\frac{1}{2}\gamma^k\|\Omega\|,x_{best}+\frac{1}{2}\gamma^k\|\Omega\|]$;\label{shrinkregion} 
                \STATE $k=k+1$;
            \ENDIF
            \STATE Project: $\rmY_t\leftarrow Proj(h_t,\tilde{\Omega})$;\label{project} 
            \STATE Sample: $(x_t,y_t)\sim \rmY_t$;
            \STATE Replace: $\mathcal{S}\leftarrow Replacing((x_t,y_t),\mathcal{S})$;
            \STATE $(x_{best},y_{best})=\arg\min\{y\colon(x,y)\in\mathcal{S}\}$;
        \ENDFOR
    \STATE {\bfseries Return:} $(x_{best},y_{best})$
    \end{algorithmic}
\end{algorithm}

The operator $\|\cdot\|$ returns a tuple represents the diameter of each dimension of the region. For instance, when $\Omega=[\omega^1_1,\omega^1_2]\times[\omega^2_1,\omega^2_2],$ we have$\|\Omega\|=(\omega^1_2-\omega^1_1, \omega^2_2-\omega^2_1).$ The projection operator $Proj(h_t,\tilde{\Omega})$ generates a random vector $\rmX_t$ with probability distribution $D_{\tilde{h}_t}:=\tilde{h}_t\slash\sP(\{x\in\Omega\colon \tilde{h}_t(x)=1\}),$ with $\tilde{h}_t(x)=1$ whenever $h_t(x)=1$ for $x\in\tilde{\Omega}.$ The sampling random vector $\rmY_t$ is induced by $\rmX_t.$ The subsequent theorem presents the upper bound of query complexity for \Algref{racecars}.
\begin{theorem}\label{racecarsconvergence}{\rm
    For \Algref{racecars} with region shrinking rate $0<\gamma<1$ and region shrinking frequency $0<\rho<1.$ Let $\epsilon>0$ and $0<\delta<1.$ When $\Omega_\epsilon$ is $\eta$-shattered by $\tilde{h}_t$ for all $t=r+1\ldots,T,$ the $(\epsilon,\delta)$-query complexity is upper bounded by 
    \begin{align*}
        \mathcal{O}\bigg(&\max\{\bigl(\frac{\gamma^{-\rho}+\gamma^{-(T-r)\rho}}{2}\lambda\eta\\
        &+(1-\lambda)\bigr)^{-1}\bigl(\frac{1}{|\Omega_\epsilon|}\ln\frac{1}{\delta}-r\bigr)+r,T\}\bigg).
        \end{align*}
    }
\end{theorem}

The condition $\gamma\in(0,1)$ ensures that the term $2p/(\gamma^{-\rho} + \gamma^{-(T-r)\rho})$ is significantly less than 1. According to Theorem \ref{racecarsconvergence}, the $(\epsilon, \delta)$-query complexity of the \Algref{racecars} is lower than that of \Algref{sequentialclassification} providing $\eta > 0$.

Theorem \ref{racecarsconvergence} establishes an upper bound on the $(\epsilon, \delta)$-query complexity that is applicable to a wide range of scenarios, assuming only that the objective function $f$ is lower-bounded. Building on this, we identify a sufficient condition for acceleration that applies to dimensionally local Holder continuity functions (Definition \ref{dimenholder}), detailed in the appendix. Within this context, the \emph{SRACOS} algorithm, which utilizes \emph{RACOS} for $Training$ phase in \Algref{sequentialclassification}, exhibit polynomial convergence \cite{hu2017sequential}. We adopt the same \emph{RACOS} approach for $Training$ sub-procedure in \Algref{racecars}, introducing "RAndomized CoordinatE Classifying And Region Shrinking" (\emph{RACE-CARS}) algorithm.

\begin{definition}[Dimensionally local Holder continuity]{\rm\label{dimenholder}
    Assume that $x_*=(x^1_*,\ldots,x^n_*)$ is the unique global minimum such that $f(x_*)=f^*.$ We call $f$ dimensionally local Holder continuity if for all $i=1,\ldots,n,$
    \begin{equation*}
        L^i_1|x^i-x^i_*|^{\beta^i_1}\le|f(x^1_*,\ldots,x^i,\ldots,x^n_*)-f^*|\le L^i_2|x^i-x^i_*|^{\beta^i_2},
    \end{equation*}
    for all $x=(x^1,\ldots,x^n)$ in the neighborhood of $X_*,$ where $\beta^i_1,$ $\beta^i_2,$ $L^i_1,$ $L^i_2$ are positive constants for $i=1,\ldots,n.$
}   
\end{definition}
 
\section{Experiments}\label{experiments}

In this section, we design experiments to test \emph{RACE-CARS} on synthetic functions, and language model tasks respectively. We use same budget to compare \emph{RACE-CARS} with a selection of DFO algorithms, including \emph{SRACOS} \cite{hu2017sequential}, zeroth-order adaptive momentum method (\emph{ZO-Adam}) \cite{chen2019zo}, differential evolution (\emph{DE}) \cite{opara2019differential} and covariance matrix adaptation evolution strategies (\emph{CMA-ES}) \cite{hansen2016cma}. All the baseline algorithms are fine-tuned, and the essential hyperparameters of \emph{RACE-CARS} can be found in Appendix.

\subsection{On Synthetic Functions}\label{synthetic}

We commence our empirical experiments with four benchmark functions: Ackley, Levy, Rastrigin and Sphere. Their analytic forms and 2-dimensional illustrations are detailed in Appendix. Characterized by extreme non-convexity and numerous local minima and saddle points—with the exception of the Sphere function—each is minimized within the boundary $\Omega = [-10,10]^n$, with a global minimum value of $0$. We choose the dimension of solution space $n$ to be $50$ and $500,$ with corresponding function evaluation budgets of $5000$ and $50000$. Notably, as indicates, the convergence of the \emph{RACE-CARS} requires only a fraction of this budget.

\begin{figure}[htb]
   \centering
   \subfigure[Ackley]{
       \includegraphics[width=0.44\linewidth]{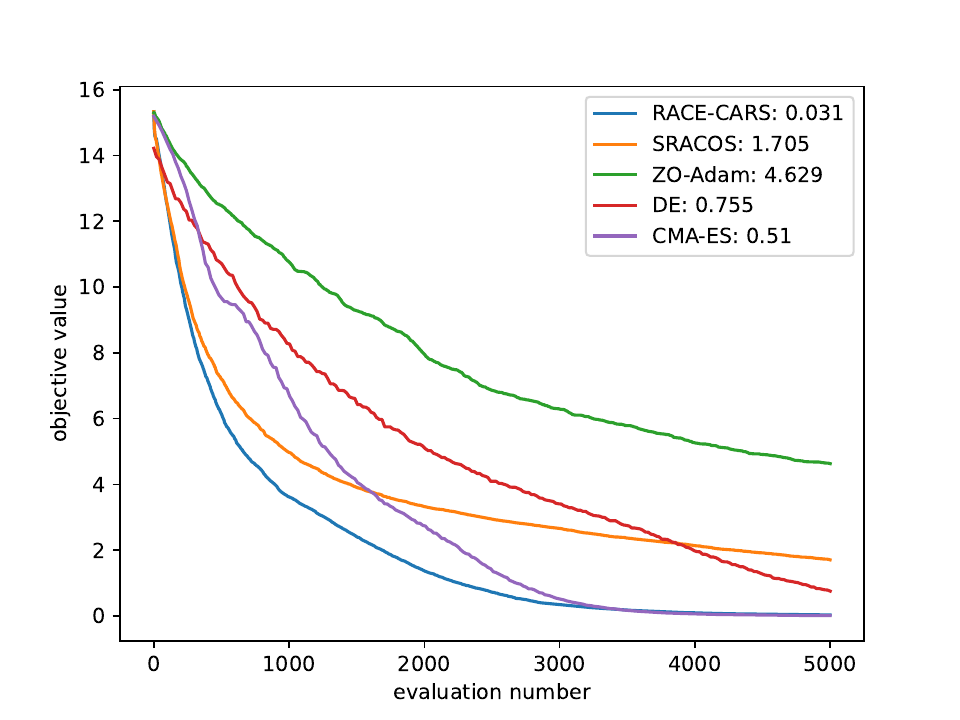}
   }
   \subfigure[Levy]{
       \includegraphics[width=0.44\linewidth]{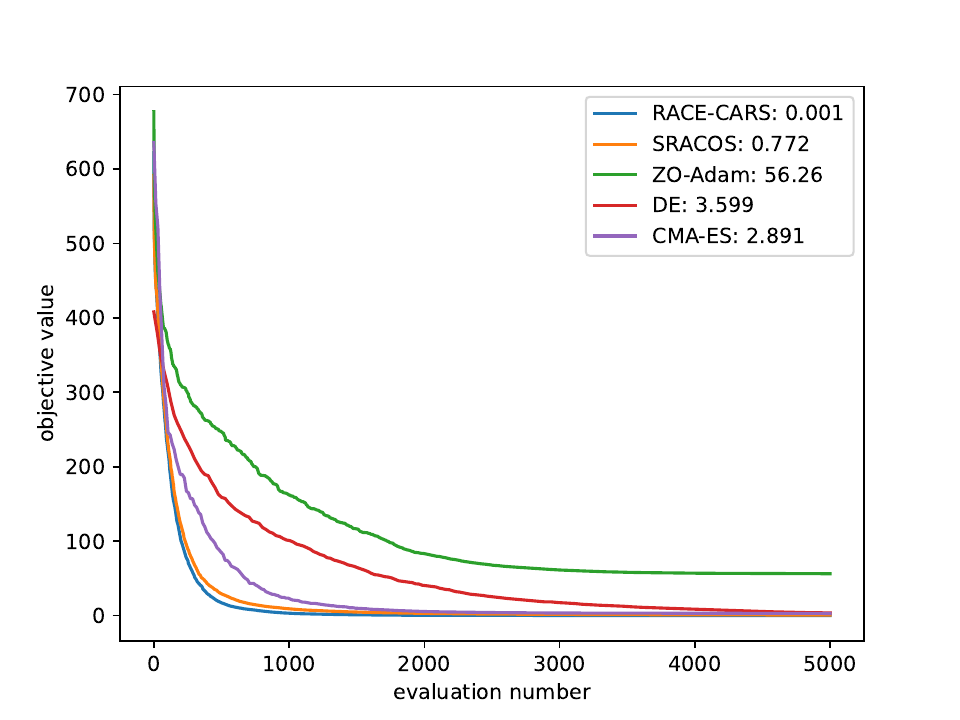}
   }
   \subfigure[Rastrigin]{
       \includegraphics[width=0.44\linewidth]{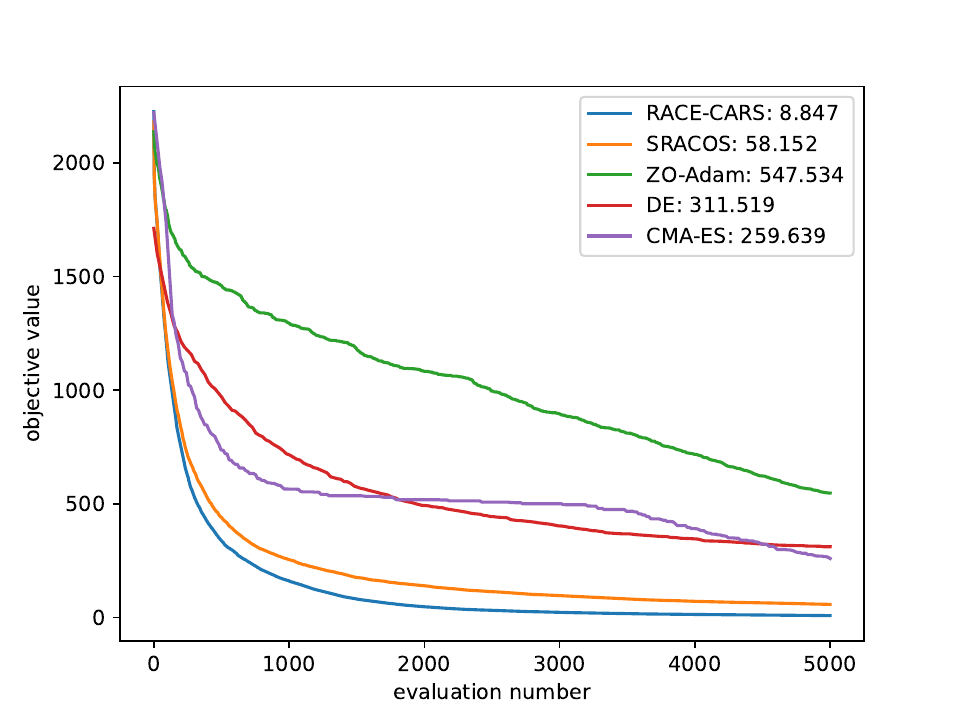}
   }
   \subfigure[Sphere]{
       \includegraphics[width=0.44\linewidth]{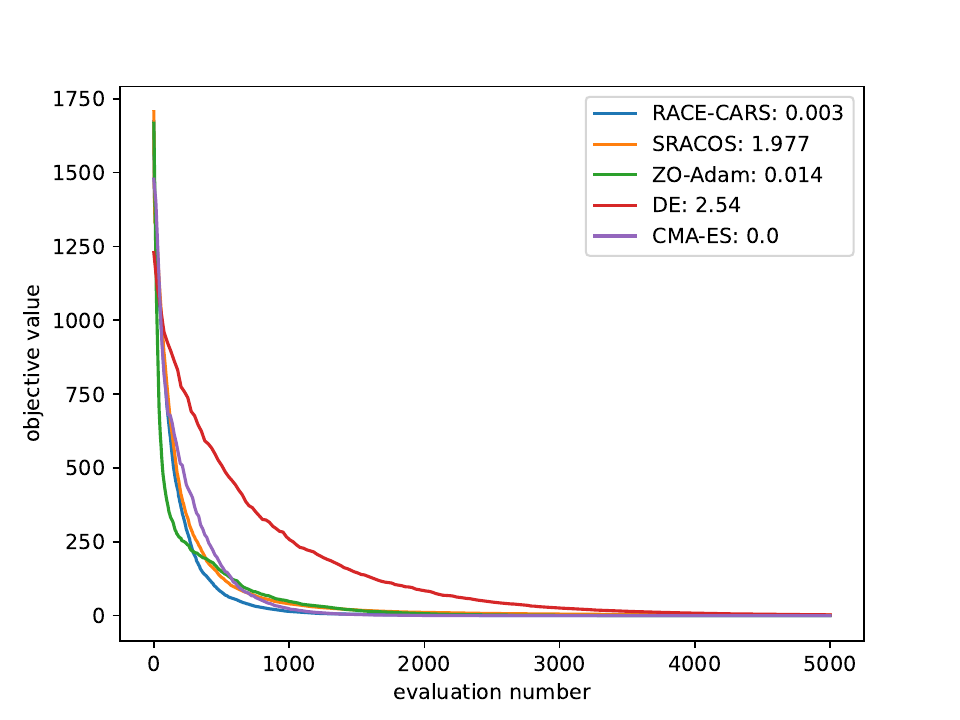}
   }
   \caption{Comparison of synthetic functions with $n=50$. }
   \label{n_50}
\end{figure}

\begin{figure}[htb]
   \centering
   \subfigure[Ackley]{
       \includegraphics[width=0.44\linewidth]{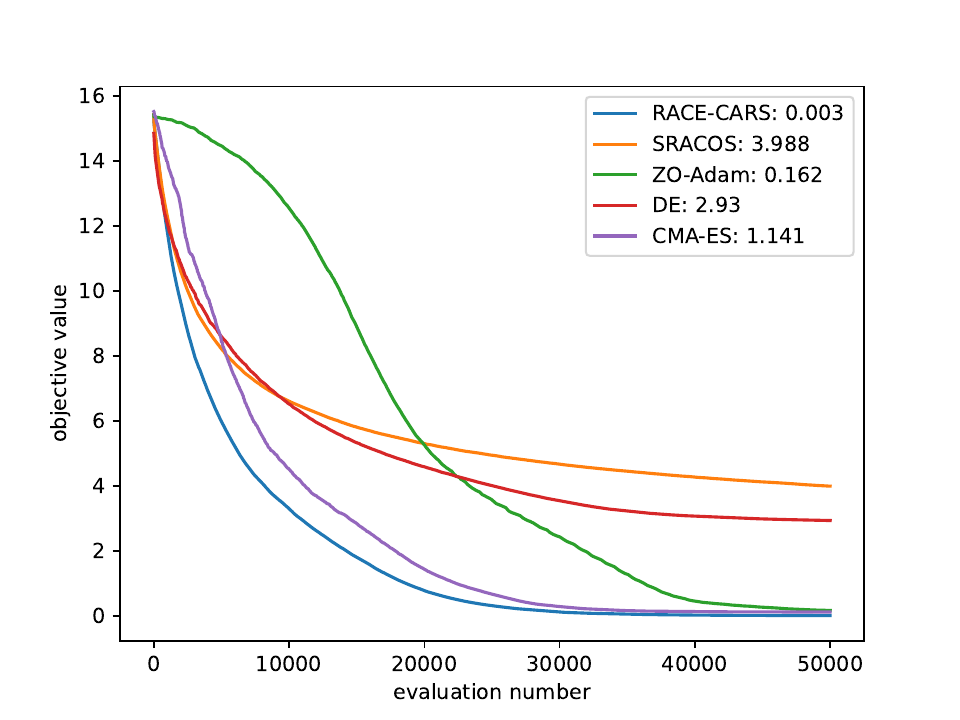}
   }
   \subfigure[Levy]{
       \includegraphics[width=0.44\linewidth]{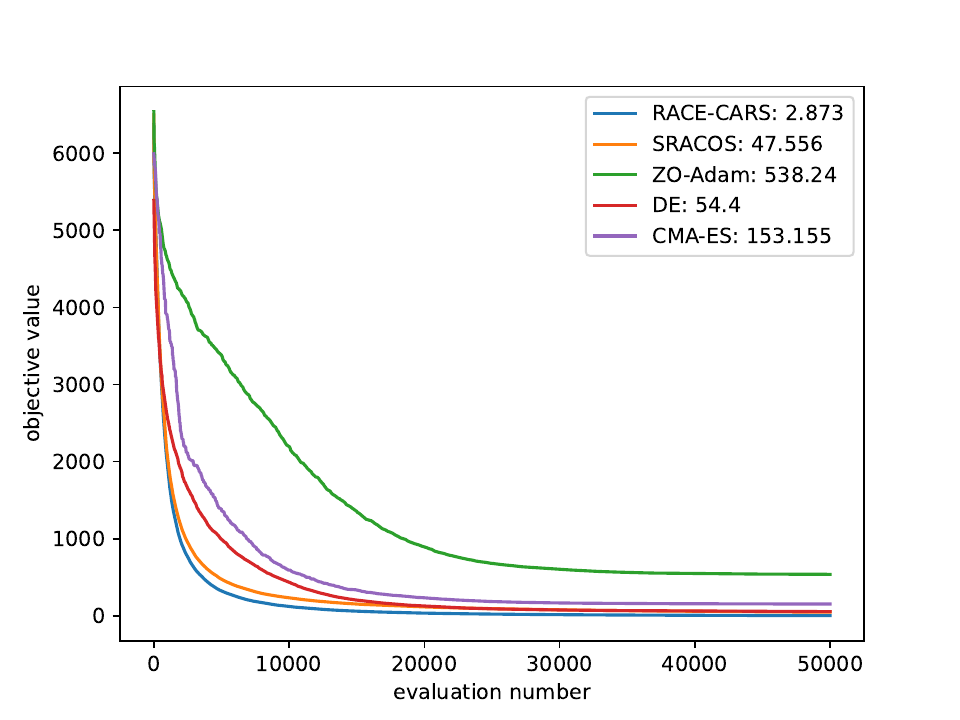}
   }\\
   \subfigure[Rastrigin]{
       \includegraphics[width=0.44\linewidth]{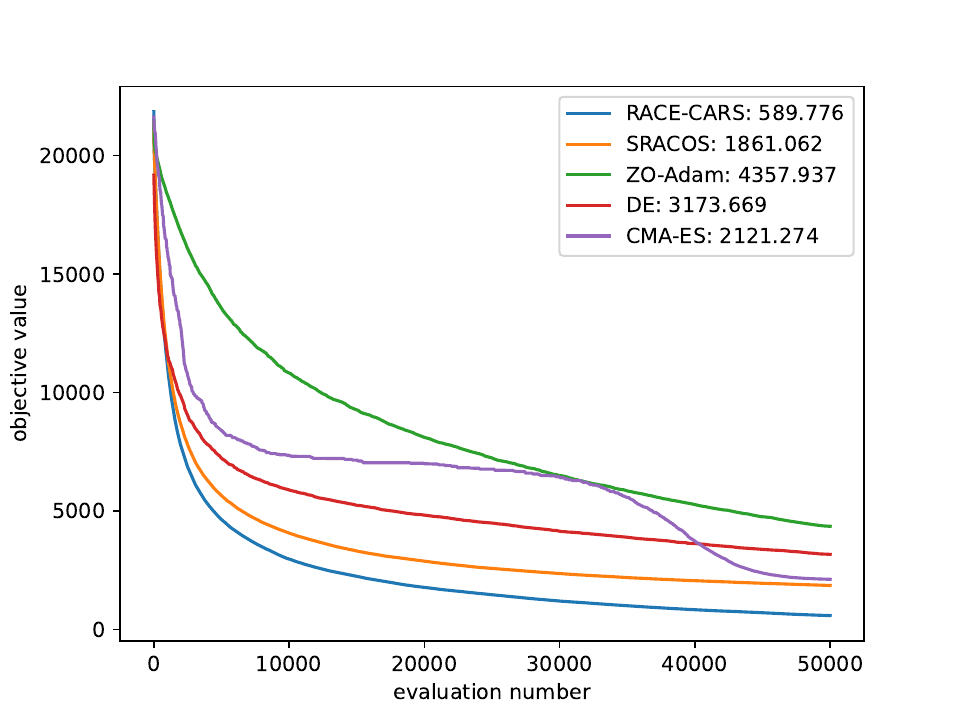}
   }
   \subfigure[Sphere]{
       \includegraphics[width=0.44\linewidth]{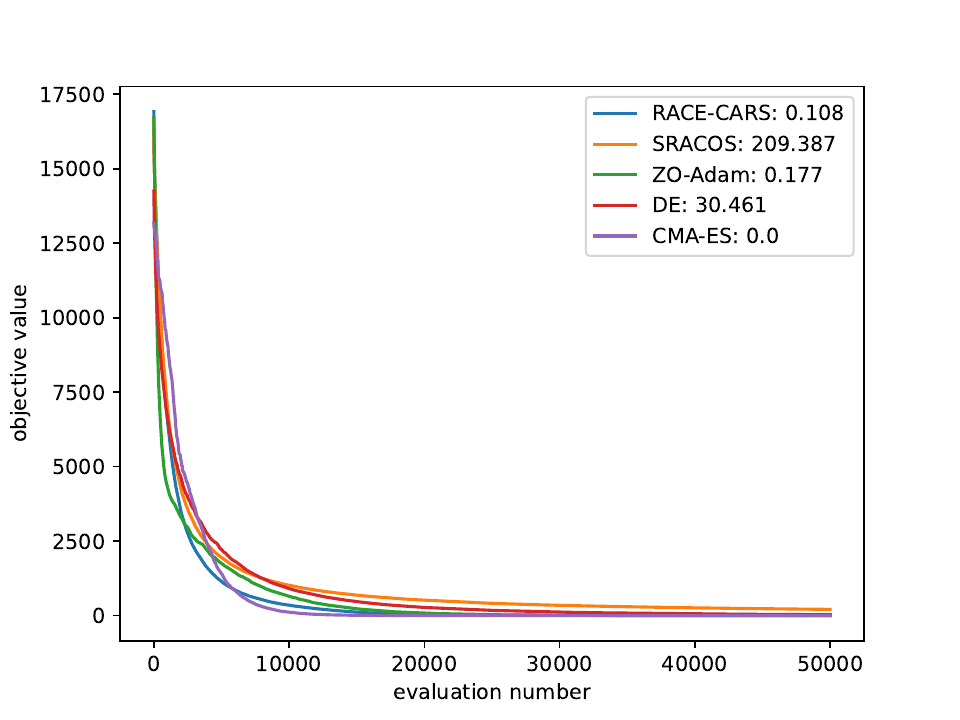}
   }
   \caption{Comparison of synthetic functions with $n=500$.}
   \label{n_500}
\end{figure}

Region shrinking rate is configured to be $\gamma=0.9$ and $0.95$, with shrinking frequency of $\rho=0.01$ and $0.001$ for $n=50,500$, respectively. Each algorithm is executed over 30 trials, and the mean convergence trajectories of the best-so-far values are depicted in \Figref{n_50} and \Figref{n_500}. The numerical values adjacent to the algorithm names in the legends represent the mean of the attained minima. It is evident that that \emph{RACE-CARS} performs the best on both convergence speed and optimal value, with a slight edge to \emph{CMA-ES} on the strongly convex Sphere function. Yet this comes at the cost of scalability due to \emph{CMA-ES}'s reliance on an $n$-dimensional covariance matrix, which is significantly more computationally intensive compared to the other algorithms.

\subsection{On Black-Box Tuning for LMaaS}

Prompt tuning for extremely large pre-trained language models (PTMs) has shown great power. PTMs such as GPT-3 \cite{brown2020language} are usually released as a service due to commercial considerations and the potential risk of misuse, allowing users to design individual prompts to query the PTMs through black-box APIs. This scenario is called Language-Model-as-a-Service (LMaaS) \cite{diao2022black,sun2022black}. In this part we follow the experiments designed by \cite{sun2022black} \footnote{Code can be found in https://github.com/txsun1997/Black-Box-Tuning}, where language understanding task is formulated as a classification task predicting for a batch of PTM-modified input texts $X$ the labels $Y$ in the PTM vocabulary, namely we need to tune the prompt such that the black-box PTM inference API $f$ takes a continuous prompt $\rvp$ satisfying $Y=f(\rvp;X).$ Moreover, to handle the high-dimensional prompt $\rvp,$ \cite{sun2022black} proposed to randomly embed the $D$-dimensional prompt $\rvp$ into a lower dimensional space $\R^d$ via random projection matrix $\rmA\in\R^{D\times d}.$ Therefore, the objective becomes:
\begin{equation*}
    \min_{z\in\mathcal{Z}}\mathcal{L}\bigl(f(\rmA z+\rvp_0;X),Y\bigr),
\end{equation*}
where $\mathcal{Z}=[-50,50]^{d}$ is the search space and $\mathcal{L}(\cdot)$ is cross entropy loss. 

In our experimental setup, we configure the search space dimension to $d=500$ and the prompt length to $50$, with RoBERTa \cite{liu2019roberta} serving as the backbone model. We evaluate performance on datasets SST-2 \cite{socher2013recursive}, Yelp polarity and AG's News  \cite{zhang2015character}, and RTE \cite{wang2018glue}. With a fixed API call budget of $T=8000,$ we pit \emph{RACE-CARS} against \emph{SRACOS} and the default DFO algorithm \emph{CMA-ES} utilized in \cite{sun2022black} \footnote{ Our choice to exclude \emph{ZO-Adam} and \emph{DE} is based on their suboptimal performance with high-dimensional nonconvex black-box functions, as demonstrated in the last section.}. 

For our tests, the shrinking rate is $\gamma=0.7$, with shrinking frequency of $\rho=0.002.$ Each algorithm is repeated 5 times independently with unique seeds. We assess the algorithms based on the mean and deviation of training loss, training accuracy, development loss and development accuracy. The SST-2 dataset results are highlighted in \Figref{sst2}, with additional findings for Yelp Polarity, AG's News, and RTE detailed in the appendix. The results indicate that \emph{RACE-CARS} consistently accelerates the convergence of \emph{SRACOS}. While \emph{CMA-ES} shows superior performance on Yelp Polarity, AG's News, and RTE, it also exhibits signs of overfitting. \emph{RACE-CARS} achieves comparable performance to \emph{CMA-ES}, despite the latter's hyperparameters being finely tuned. Notably, the hyperparameters for \emph{RACE-CARS} were empirically adjusted based on the SST-2 dataset and then applied to the other three datasets without further tuning.

\begin{figure}[htb]
    \centering
    \subfigure[Training loss]{
        \includegraphics[width=0.44\linewidth]{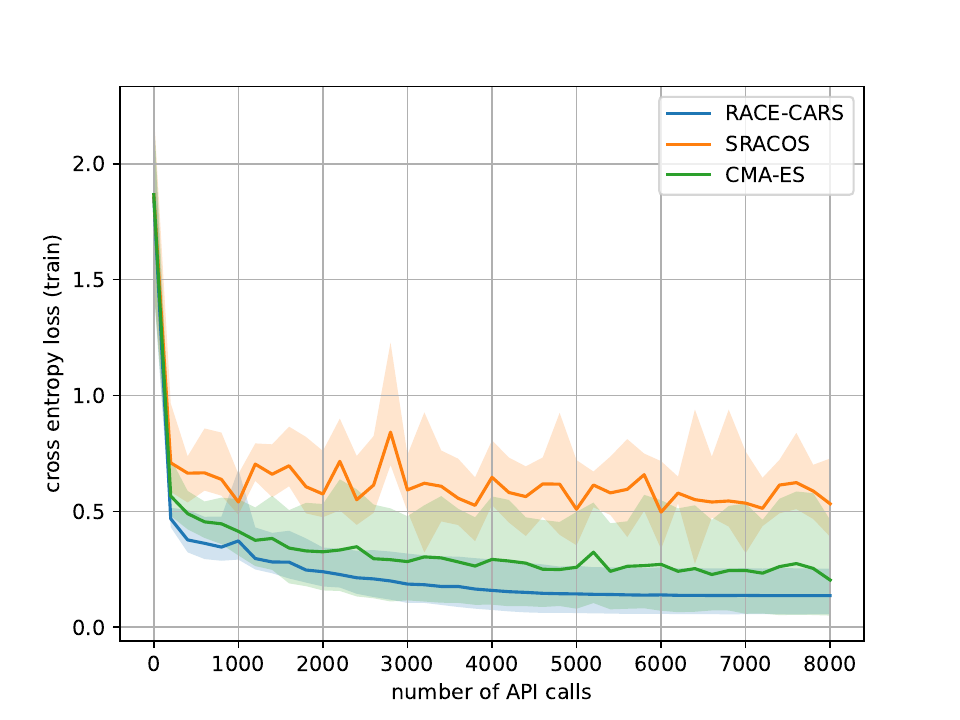}
    }
    \subfigure[Training accuracy]{
        \includegraphics[width=0.44\linewidth]{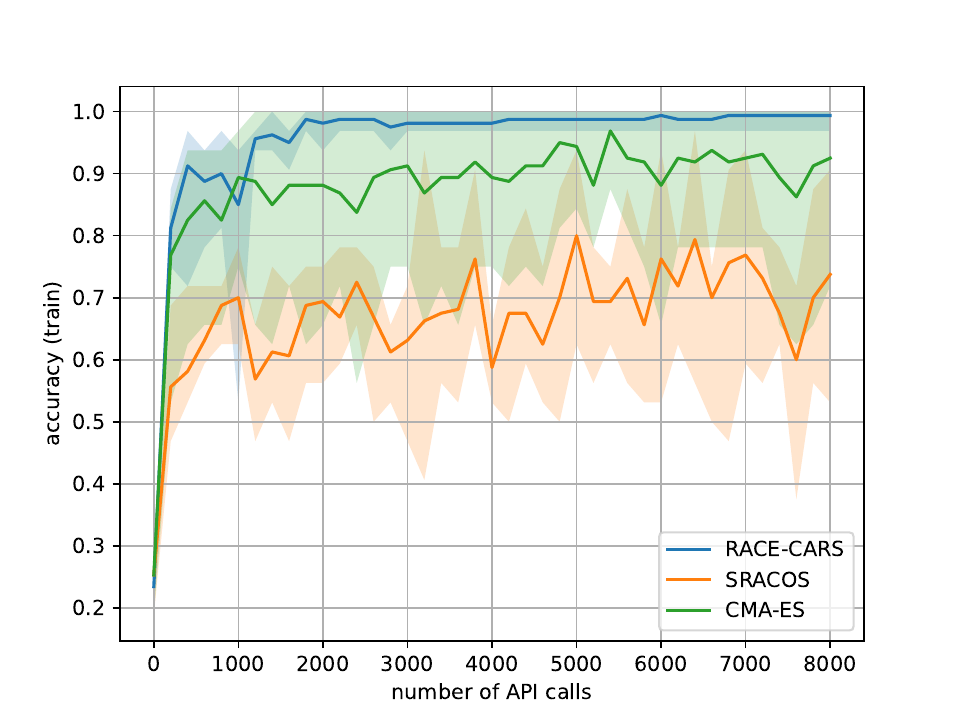}
    }\\
    \subfigure[Development loss]{
        \includegraphics[width=0.44\linewidth]{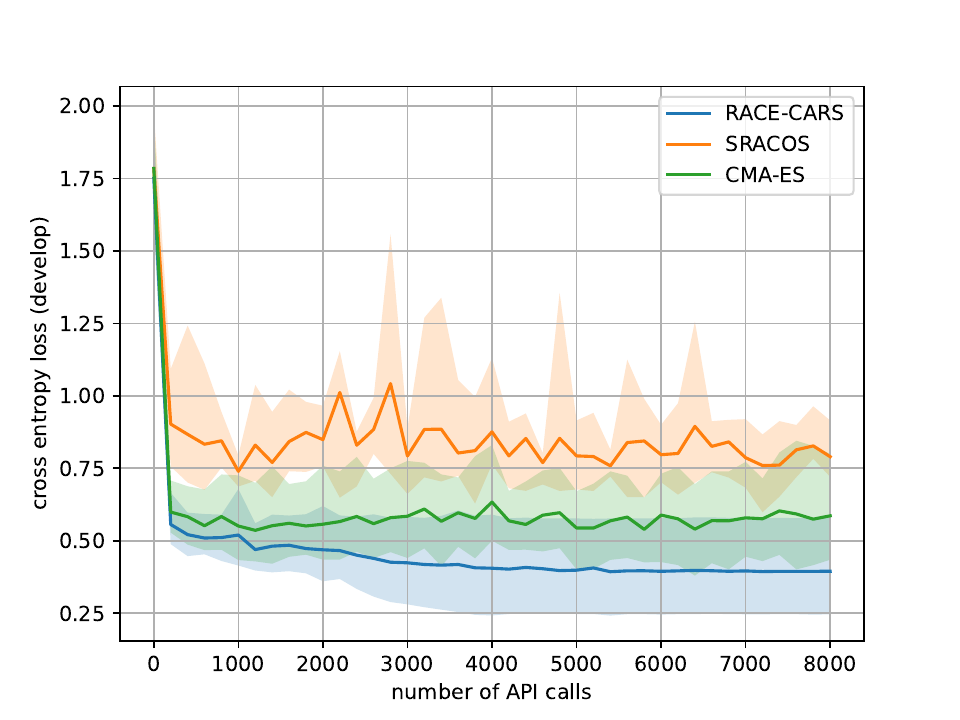}
    }
    \subfigure[Development accuracy]{
        \includegraphics[width=0.44\linewidth]{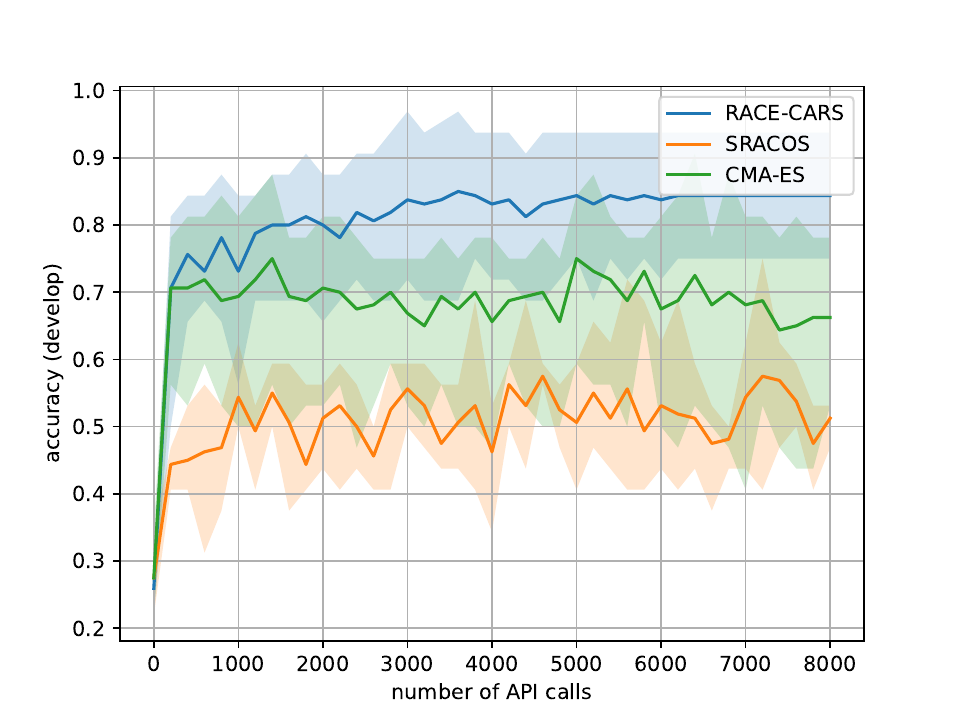}
    }
    \caption{Comparisons on SST-2}
    \label{sst2}
\end{figure}

\section{Discussion}\label{discussion}
\subsection{Beyond continuity}

\begin{itemize}
   \item [{\bf (i)}] {\bf For discontinuous objective functions.}
   
   Dimensionally local Holder continuity in Definition \ref{dimenholder}, imposes certain constraints on the objective through a set of continuous envelopes, whereas the objective is not supposed to be continuous. Beyond the continuous cases discussed in the previous section, \emph{RACE-CARS} remains applicable to discontinuous objectives as well. Refer to appendix for a comprehensive understanding.
   \item [{\bf (ii)}] {\bf For discrete optimization.}
   
   Similar to \emph{SRACOS} algorithm, \emph{RACE-CARS} retains the capability to tackle discrete optimization problems. The convergence theorems, presented in Theorem \ref{sracosconvergence} and Theorem \ref{racecarsconvergence}, encompass this situation by altering the measure of probability space to be for example, induced by counting measure. We extend experiments on mixed-integer programming problems, substantiating the acceleration of \emph{RACE-CARS} empirically. See appendix for details.
\end{itemize}

\subsection{On the Concept Hypothesis-Target Shattering}

The concept hypothesis-target shattering, central to our discussion, draws inspiration from the established learning theory notion of \emph{shatter} and its deep ties to the Vapnik-Chervonenkis (VC) theory \cite{vapnik1998statistical}. At the heart of VC theory lies the VC dimension, a measure of a hypothesis family's capacity to distinguish among data points based on their labels. Specifically, for a collection of data points with binary labels, $S$, we say a subset $S' \subseteq S$ is shattered by a hypothesis family $\mathcal{H}$ if there exists a hypothesis $h \in \mathcal{H}$ that perfectly aligns with the labels of points in $S'$ and contrasts with those outside:
\begin{equation*}
    h(x)=
    \begin{cases}
        1, &x\in S^\prime\\
        0, &x\notin S^\prime.
    \end{cases}
\end{equation*}
The shattering coefficient, $\mathcal{S}(\mathcal{H}, n)$, signifies the variety of point-label combinations that $\mathcal{H}$ can produce for $n$ points. The VC dimension is then defined as $VC(\mathcal{H}) := \{\sup{n : \mathcal{S}(\mathcal{H}, n) = 2^n}\}$, indicating the maximum number of points that can be distinctly labeled by $\mathcal{H}$.

In the context of classification-based DFO, we represent the target-representative capability of a family of hypotheses through hypothesis-target shattering, measuring the overlap of hypothesis's active region and the target. Therefore the quintessence of algorithm design hinges on maximizing this quantity. discerning the target-representative capacity within the intricate landscape of \emph{nonconvex}, \emph{black-box} optimization problems is nontrivial. Nonetheless, the target-representative capability of hypotheses family generated by \emph{RACOS}, although proved under the previous framework, empirically suggests sufficient efficacy in scenarios where the objective function exhibits locally Holder continuouity. Looking ahead, altering the $Training$ and $Replacing$ sub-procedures inherited from \emph{SRACOS}, which may ideally lead to a bigger shattering rate and maintain the easy-to-sample characterization, will be another extension direction of the current study.

\subsection{Ablation Experiments}

While it's an appealing goal to develop a universally effective DFO algorithm for black-box functions without the need for hyperparameter tuning, this remains an unrealistic aspiration. Our proposed algorithm, \emph{RACE-CARS}, introduces two hyperparameters: shrinking-rate $\gamma$ and shrinking-frequency $\rho.$ For an $n$-dimensional optimization problem, we call $\gamma^{n\rho}$ shrinking factor of \emph{RACE-CARS}. We take Ackley for a case study, design ablation experiments on the two hyperparameters of \emph{RACE-CARS} to reveal the mechanism. We stipulate that we do not aim to identify a optimal combination of hyperparameters for maximizing the overlap with the target hypothesis.  Instead, our aim is to provide empirical guidance for tuning these hyperparameters effectively. For further details, the reader is directed to the appendix.

\section{Conclusion}

In this paper, we refine the framework of classification-based DFO as a stochastic process, and propose a novel learning concept named hypothesis-target shattering rate. Our research delves into the convergence properties of sequential-mode classification-based DFO algorithms and provides a fresh perspective on their query complexity upper bound. Delighted by the computational complexity upper bound under the new framework, we propose a theoretically grounded region-shrinking technique to accelerate the convergence. In empirical analysis, we study the scalability performance of \emph{RACE-CARS} on both synthetic functions and black-box tuning for LMaaS, showing its superiority over \emph{SRACOS}.

\bibliography{aaai25}

\onecolumn
 
\newpage

\appendix
\section{Appendix}
\subsection{Synthetic functions}\label{syn}

\begin{itemize}
    \item Ackley:
    \begin{align*}
        f(x)=&-20\exp(-0.2\sqrt{\sum_{i=1}^{n}(x_i-0.2)^2\slash n})\\
        &-\exp(\sum_{i=1}^{n}\cos2\pi x_i\slash n)+e+20.
    \end{align*}
    \item Levy:
    \begin{align*}
        f(x)=&\sin^2(\pi\omega_1)+\sum_{i=1}^{n-1}(\omega_i-1)^2\bigl(1+10\sin^2(\pi\omega_i+1)\bigr)\\
        &+(\omega_n-1)^2\bigl(1+\sin^2(2\pi\omega_n)\bigr)
    \end{align*}
    where $\omega_i=1+\frac{x_i-1}{4}.$
    \item Rastrigin:
    \begin{equation*}
        f(x)=10n+\sum_{i=1}^{n}\bigl(x_i^2-10\cos(2\pi x_i)\bigr).
    \end{equation*}
    \item Sphere:
    \begin{equation*}
        f(x)=\sum_{i=1}^{n}(x_i-0.2)^2.
    \end{equation*}
\end{itemize}
\begin{figure}[htb]
    \centering
    \subfigure[Ackley]{
        \includegraphics[width=0.345\linewidth]{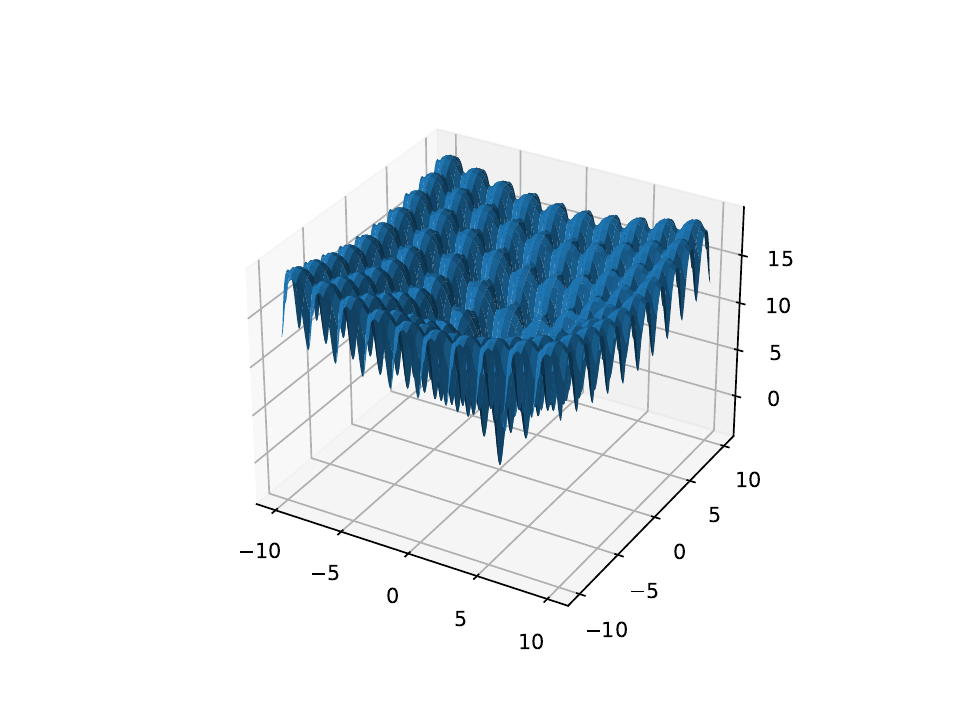}
    }
    \subfigure[Levy]{
        \includegraphics[width=0.345\linewidth]{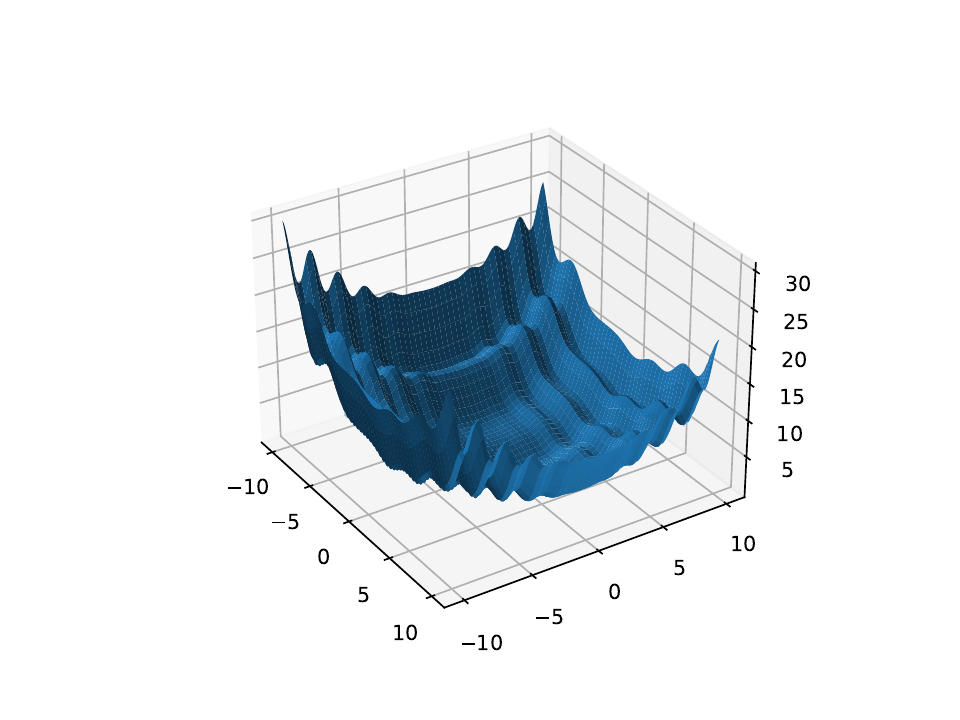}
    }\\
    \subfigure[Rastrigin]{
        \includegraphics[width=0.345\linewidth]{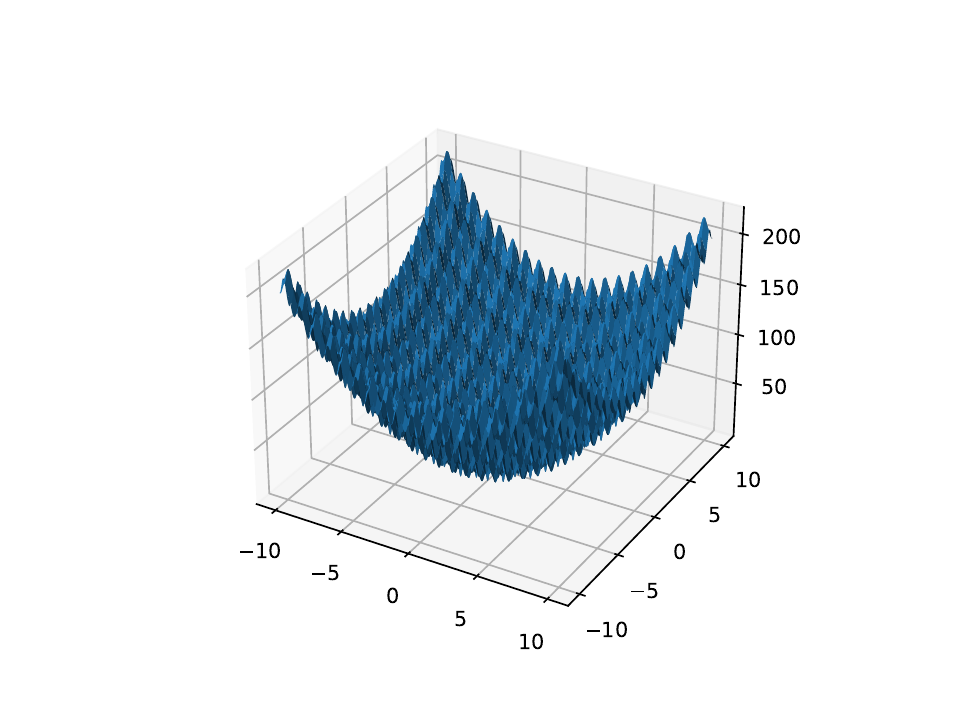}
    }
    \subfigure[Sphere]{
        \includegraphics[width=0.345\linewidth]{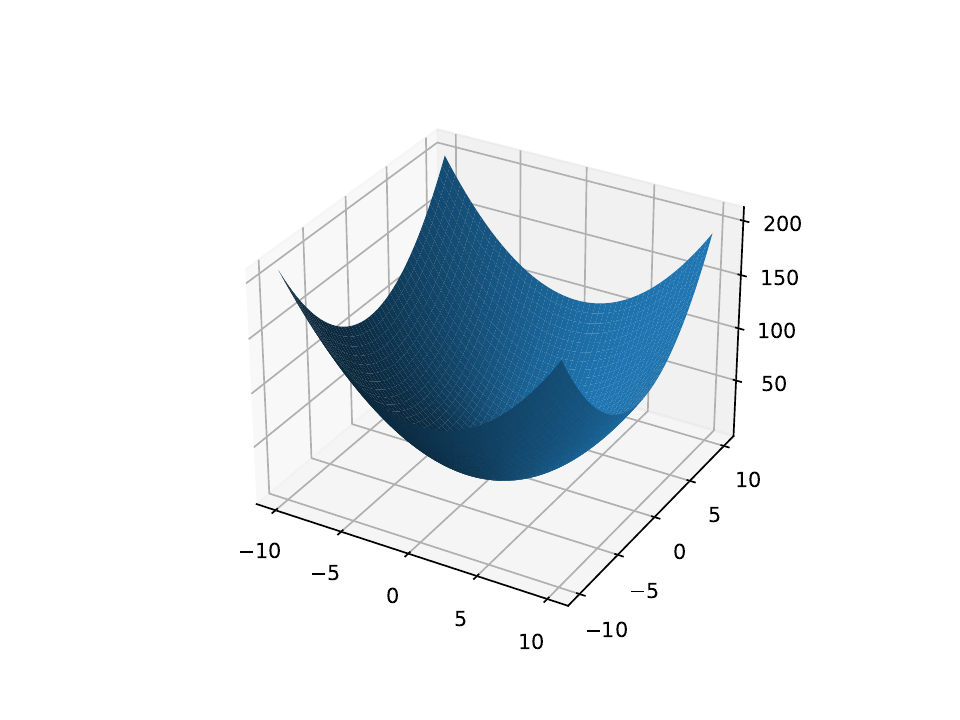}
    }
    \caption{Synthetic functions with $n=2$.}
    \label{benchmarks}
\end{figure}

\subsection{Beyond continuity} \label{appendixBeyond}
       
\subsubsection{For discontinuous objective functions}\label{appendixdiscontinuousopt}
\begin{figure}[htb]
    \centering
    \subfigure[Ackley]{
        \includegraphics[width=0.44\linewidth]{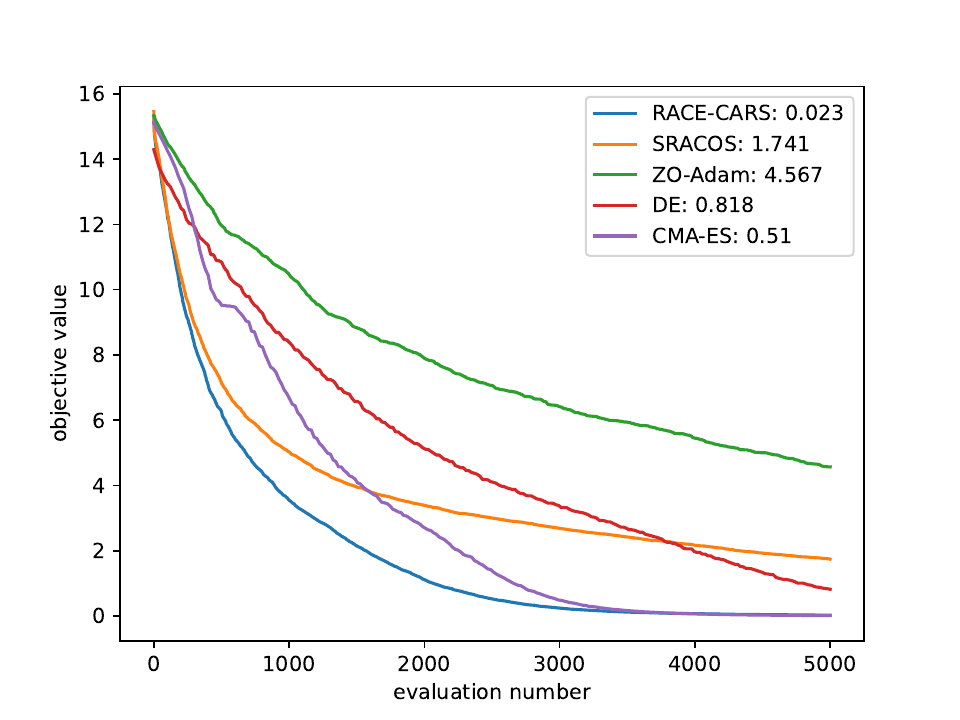}
    }
    \subfigure[Levy]{
        \includegraphics[width=0.44\linewidth]{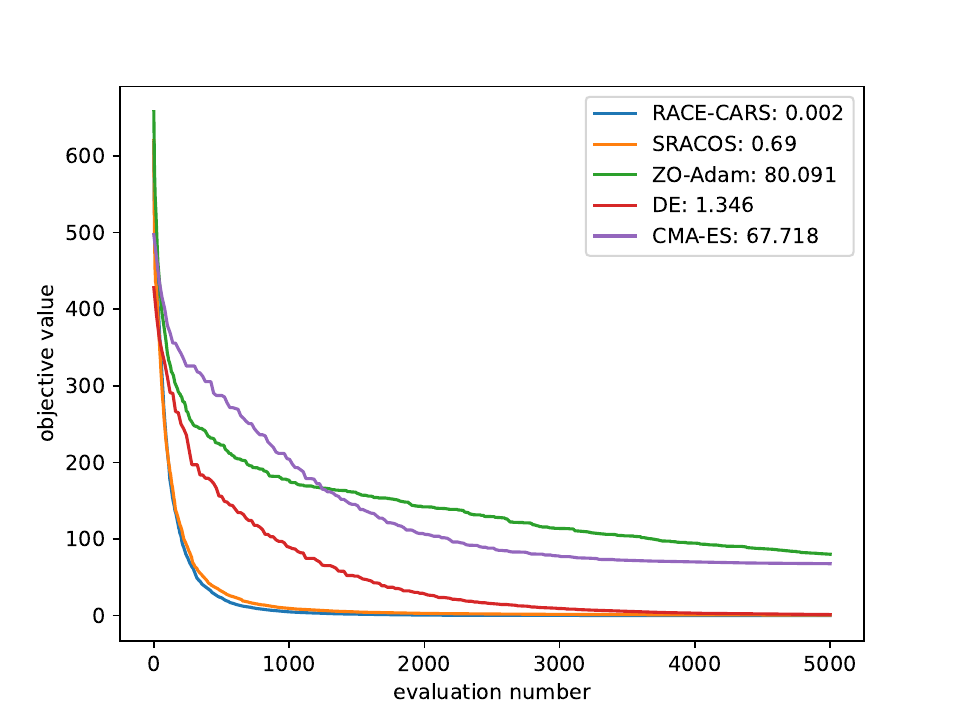}
    }\\
    \subfigure[Rastrigin]{
        \includegraphics[width=0.44\linewidth]{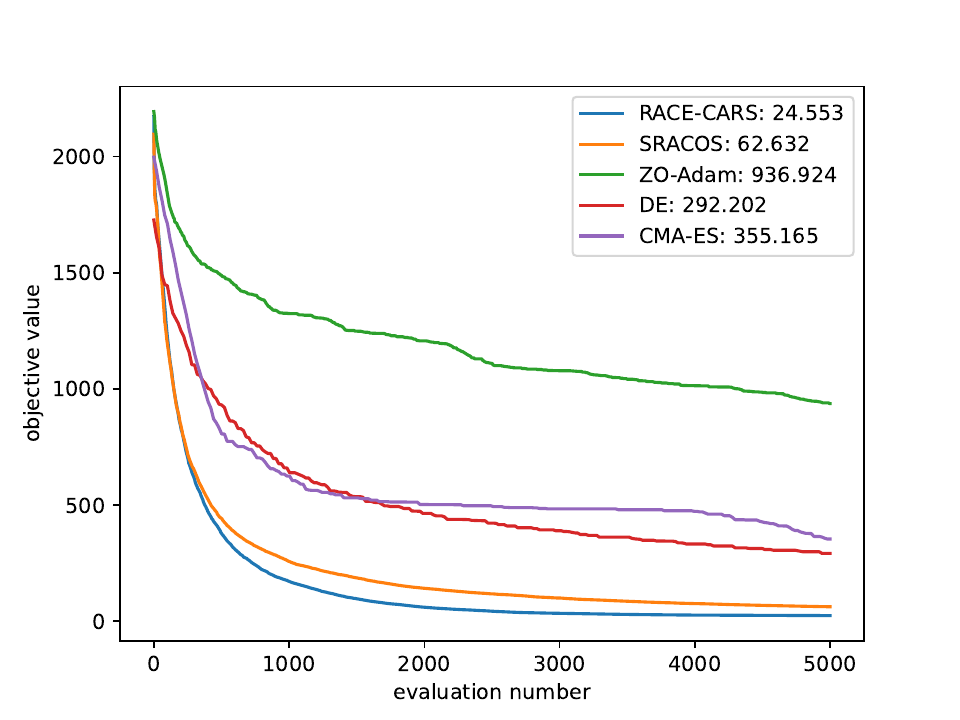}
    }
    \subfigure[Sphere]{
        \includegraphics[width=0.44\linewidth]{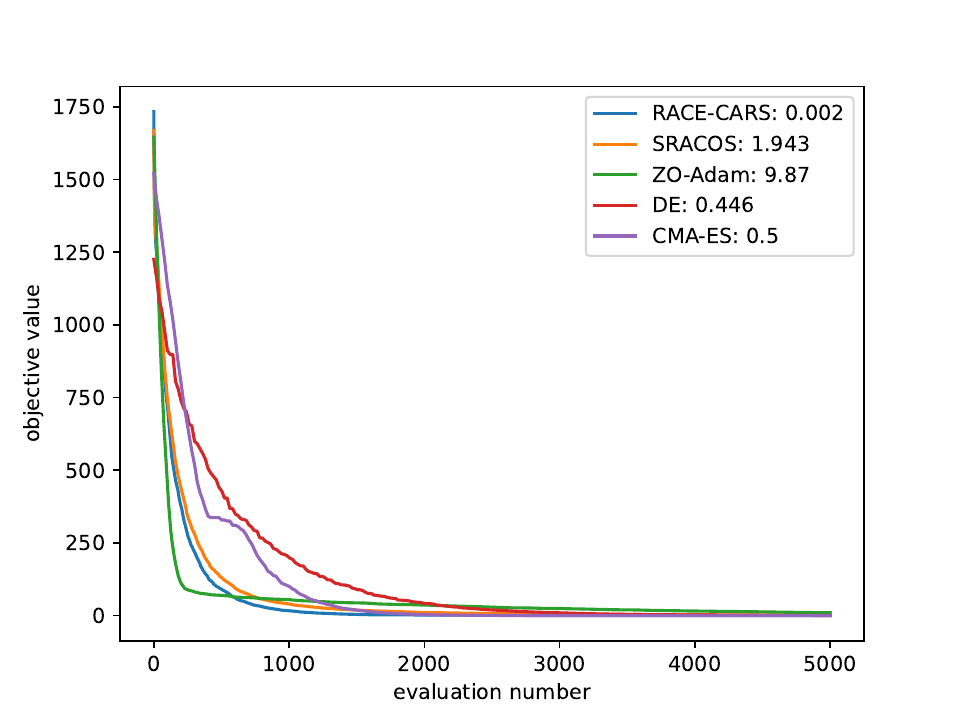}
    }
    \caption{Comparison on discontinuous objectives}
    \label{discontinuous}
\end{figure}

We design experiments on discontinuous objective functions by adding random perturbation to synthetic functions. For example, the perturbation is set to be:
\begin{equation*}
   P(x)=\sum_{i=1}^{m}\epsilon_i\cdot\delta_{\mathcal{B}(x_i,0.5)}(x).
\end{equation*}
$\mathcal{B}(x_i,0.5)$ is the open ball centering at $x_i$ with radius equals to 0.5, with $x_i,$ $i=1,\ldots,m,$ randomly generated within the solution region. $\delta_{\mathcal{B}(x_i,0.5)}(x)$ is an indicator function, equaling to 1 when $x\in\mathcal{B}(x_i,0.5)$ otherwise 0. The perturbations $\epsilon_i$ are uniformly sampled from $[0, 1]$ for every single ball center $x_i,$ $i=1,\ldots,m.$ The objective function is set to be 
\begin{equation*}
    \tilde{f}(x):=f(x)+P(x),
\end{equation*}
which is lower semi-continuous. We use the same settings as in the body sections with dimension $n=50,$ the perturbation size $m=5n$ and budget $T=100n.$ Similarly, region shrinking rate is set to be $\gamma=0.9$ and region shrinking frequency is $\rho=0.01.$ Each of the algorithm is repeated 30 runs and the mean convergence trajectories of the best-so-far values are presented in \Figref{discontinuous}. The numbers attached to the algorithm names in the legend of figures are the mean value of obtained minimum. It can be observed that the acceleration of \emph{RACE-CARS} to \emph{SRACOS} is still valid. Comparing with baselines, \emph{RACE-CARS} performs the best on both convergence, and obtain the best optimal value. As we anticipated, the performance of \emph{SRACOS} and \emph{RACE-CARS} are almost impervious to discontinuity, whereas the other three baselines, whose convergence relies on the continuity, suffers from oscillation or early-stopping to different extent. 

\subsubsection{For discrete optimization}\label{appendixdiscreteopt}

In order to transfer \emph{RACE-CARS} to discrete optimization, $Training$ and $Projection$ sub-procedures in \Algref{racecars} should be modified. In all cases, we employ the discrete version of \emph{RACOS} for $Training$ \cite{yu2016derivative}. Furthermore, we presume counting measure $\#$ as the inducing measure of probability space $(\Omega, \mathcal{F}, \mathbb{P}),$ where $\mathbb{P}(B):=\#(B)/\#(\Omega)$ for all $B\in\mathcal{F}.$ The $Projection$ is therefore similar only to set the operator $\|\cdot\|$ return the count of each dimension of the region. 

\begin{figure}[htb]
   \centering
   \subfigure[$n_1=n_2=50$]{
       \includegraphics[width=0.44\linewidth]{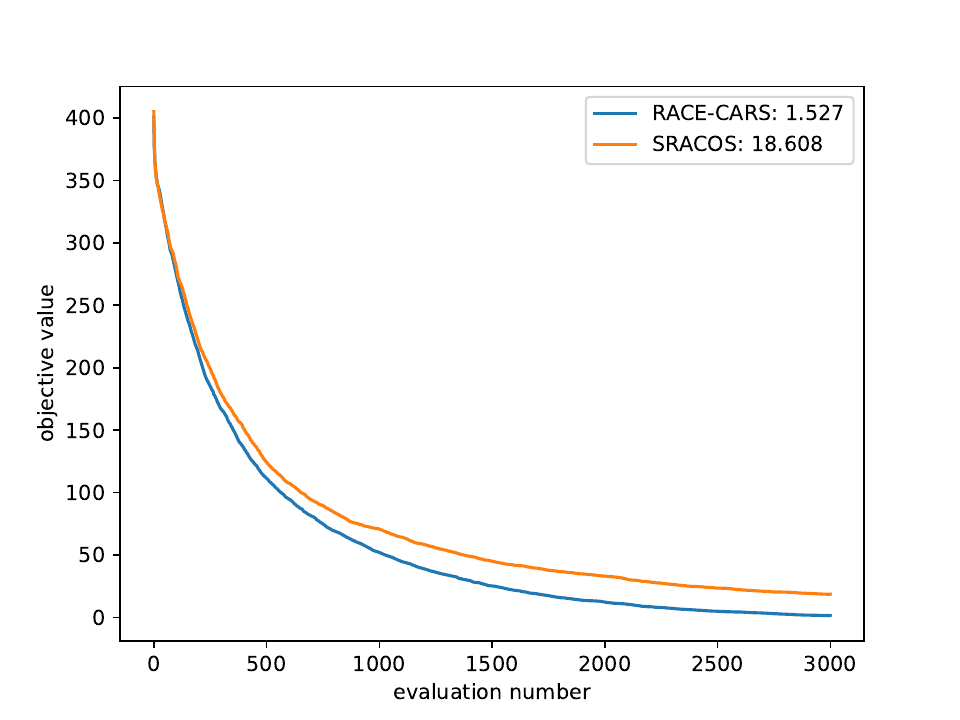}
   }
   \subfigure[$n_1=n_2=250$]{
       \includegraphics[width=0.44\linewidth]{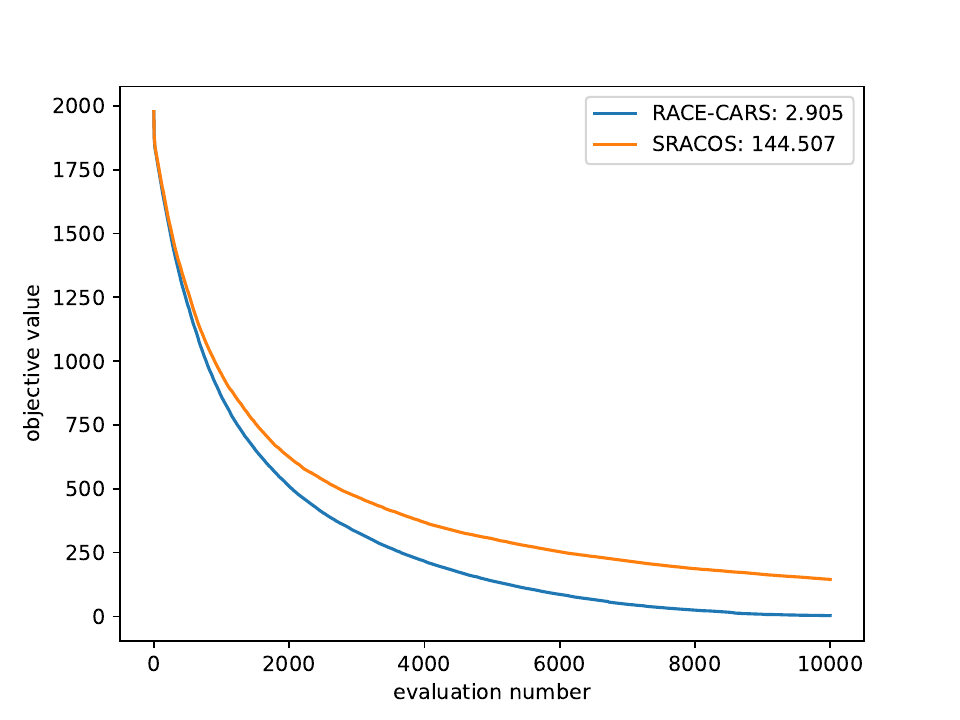}
   }
   \caption{Mixed-integer programming.}
   \label{discreteresults}
\end{figure}

We design experiments on the following formulation:
\begin{align}
   \min~&f(x, y)\label{discretemodel}\\
   s.t.~&x\in\Omega_c\notag\\
        &y\in\Omega_d\notag,
\end{align}
where $\Omega_c$ is the continuous solution subspace and $\Omega_d$ is discrete. \Eqref{discretemodel} encompasses a wide range of continuous, discrete and mixed-integer programming problems. In our experiments, we specify \eqref{discretemodel} as a mixed-integer programming problem:
\begin{align*}
   \min~&Ackley(x)+L^Tabs(y)\\
   s.t.~&x\in[-1,1]^{n_1}\\
        &y\in\{-10,-9,\ldots,9,10\}^{n_2},
\end{align*}
where $L\in\mathbb{R}^{n_2}$ is sampled uniformly from $[1, 2]^{n_2},$ thus the global optimal value is 0. We choose the dimension of solution space as $n_1=n_2=50$ and $250,$ the budget of function evaluation is set to be $3000$ and $10000$ respectively. Region shrinking rate is set to be $\gamma=0.95$ and region shrinking frequency is $\rho=0.01,$ $0.005$ respectively. Each of the algorithm is repeated 30 runs and the convergence trajectories of mean of the best-so-far value are presented in Figure \ref{discreteresults}. As results show, \emph{RACE-CARS} maintains acceleration to \emph{SRACOS} in discrete situation.

\subsection{Ablation experiments}\label{appendixablation}
\begin{itemize}
    \item [\textbf{(i)}] \textbf{Relationship between shrinking frequency $\mathbf{\rho}$ and dimension $\mathbf{n}$.}

    \begin{minipage}[htbp]{.54\linewidth}
        For Ackley on $\Omega=[-10,10]^n,$ we fix shrinking rate $\rho=0.95$ and compare the performance of \emph{RACE-CARS} between different shrinking frequency $\rho$ and dimension $n.$ The shrinking frequencies $\rho$ ranges from $0.002$ to $0.2$ and dimension $n$ ranges from $50$ to $500.$ The function calls budget is set to be $T=30n$ for fair. Experiments are repeated 5 times for each hyperparameter and results are recorded in Table \ref{table::1} and the normalized results is the heatmap figure in the right. The black curve represents the trajectory of best shrinking frequency with respect to dimension. The horizontal axis is dimension and the vertical axis is shrinking frequency. Results indicate the best $\rho$ is in reverse proportion to $n,$ therefore maintaining $n\rho$ as constant is preferred.
    \end{minipage}
    \begin{minipage}[htbp]{.45\linewidth}
        \centering
        \includegraphics[width=0.9\linewidth]{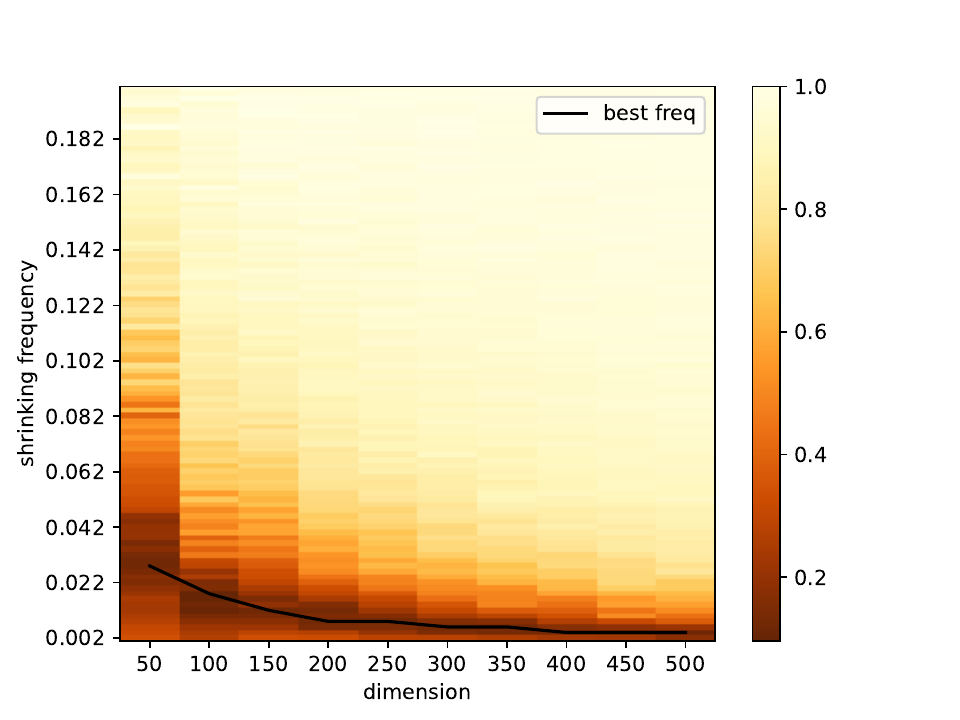}
    \end{minipage}
   
    \item [\textbf{(ii)}] \textbf{Relationship between shrinking factor $\mathbf{\gamma^{n\rho}}$ and dimension $\mathbf{n}$ of solution space.}
    
    For Ackley on $\Omega=[-r,r]^n,$ we compare the performance of \emph{RACE-CARS} between different shrinking factors and radius $r.$ Different shrinking factors are generated by varying shrinking rate $\gamma$ and dimension times shrinking frequency $n\rho.$ We design experiments on 4 different dimensions $n$ with 4 radii $r.$ The function calls budget is set to be $T=30n.$ Experiments are repeated 5 times for each hyperparameter and results are presented in heatmap format in Figure \ref{factordim}. According to the results, the best shrinking factor is insensitive to the variation of dimension. Considering that the best $n\rho$ maintains constant as $n$ varying, slightly variation of the corresponding best $\gamma$ is preferred. This observation is in line with what we anticipated as in section \ref{experiments}.
           
    \begin{figure}[htb]
        \centering
        \subfigure[Radius $r=1$]{
            \includegraphics[width=0.22\linewidth]{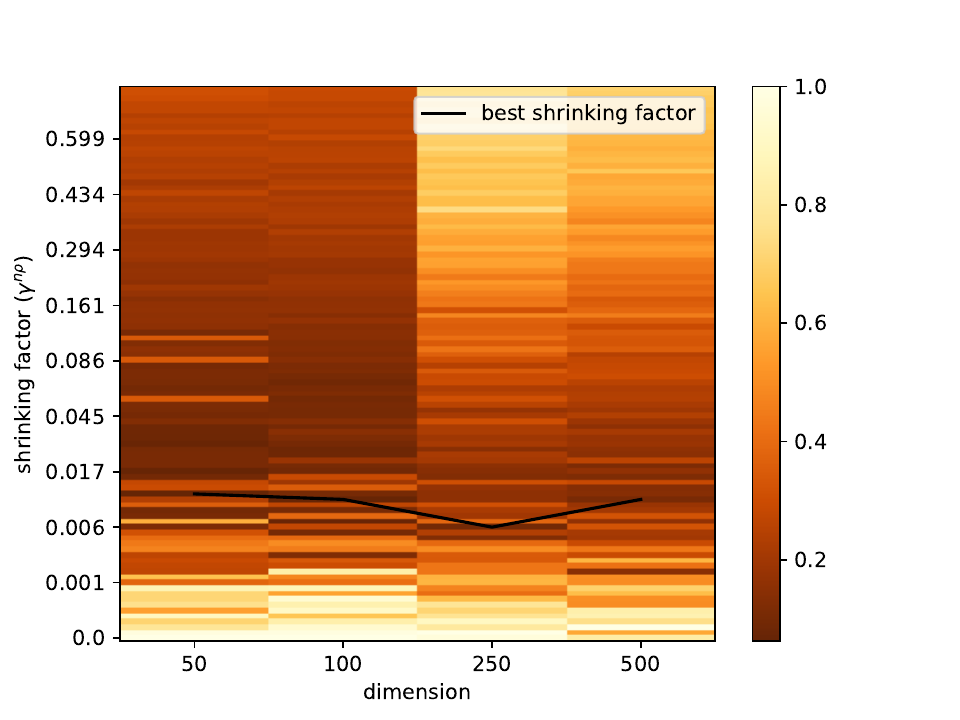}
        }
        \subfigure[Radius $r=5$]{
            \includegraphics[width=0.22\linewidth]{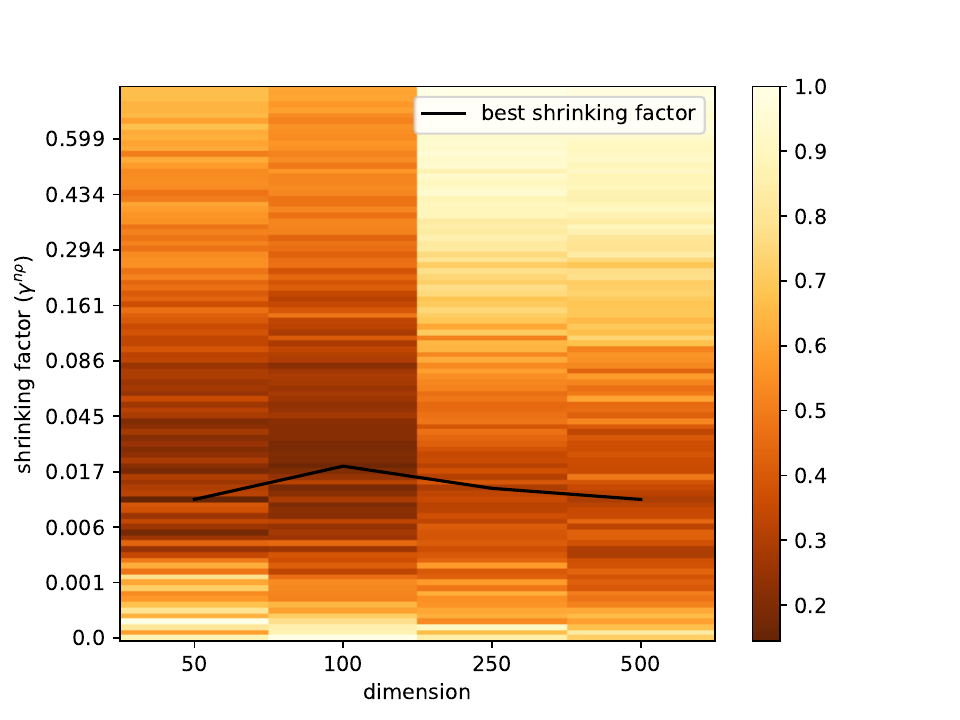}
        }
        \subfigure[Radius $r=10$]{
            \includegraphics[width=0.22\linewidth]{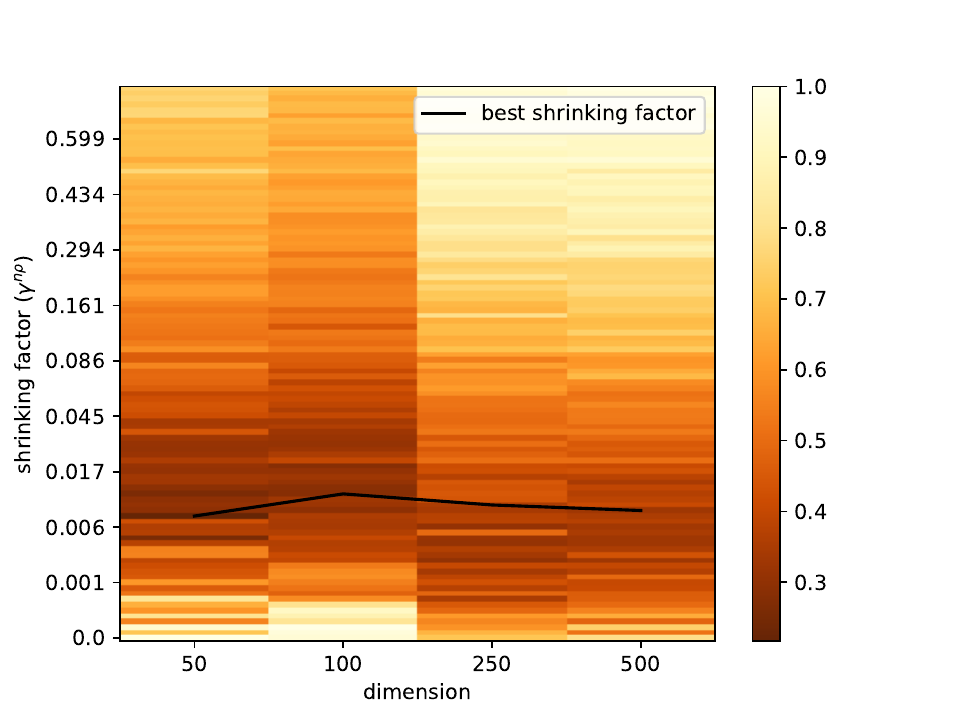}
        }
        \subfigure[Radius $r=25$]{
            \includegraphics[width=0.22\linewidth]{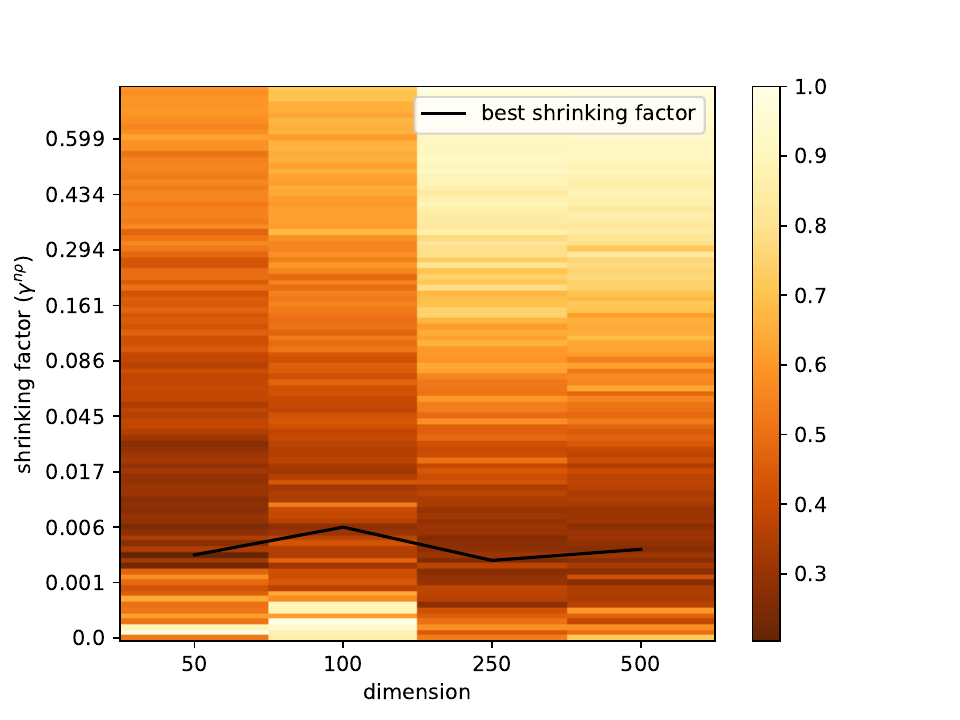}
        }
        \caption{\textbf{Comparison of shrinking factor $\mathbf{\gamma^{n\rho}}$ and dimension $\mathbf{n}$ of the solution space $\mathbf{\Omega=[-r,r]^n}$. }{\small
        Results of different solution space radius are presented in each subfigure respectively. In each subfigure, the horizontal axis is the dimension and the vertical axis is shrinking factor. Each pixel represents the heat of y-wise normalized mean function value at the $30n$ step. The black curve is the best shrinking factor in each dimension.
        }}
        \label{factordim}
    \end{figure}
    \item [\textbf{(iii)}] \textbf{Relationship between shrinking factor $\mathbf{\gamma^{n\rho}}$ and radius $\mathbf{r}$ of solution space.}
    
    For Ackley on $\Omega=[-r,r]^n,$ we compare the performance of \emph{RACE-CARS} between different shrinking factors and radius $r.$ Different shrinking factors are generated by varying shrinking rate $\gamma$ and dimension times shrinking frequency $n\rho.$ We design experiments on 4 different radii $r$ with 4 dimensions $n.$ The function calls budget is set to be $T=30n.$ Experiments are repeated 5 times for each hyperparameter and results are presented in heatmap format in Figure \ref{factorradius}. According to the results, the best shrinking factor $\gamma^{n\rho}$ should be decreased as radius $r$ increases.
    
    \begin{figure}[htb]
        \centering
        \subfigure[Dimension $n=50$]{
            \includegraphics[width=0.22\linewidth]{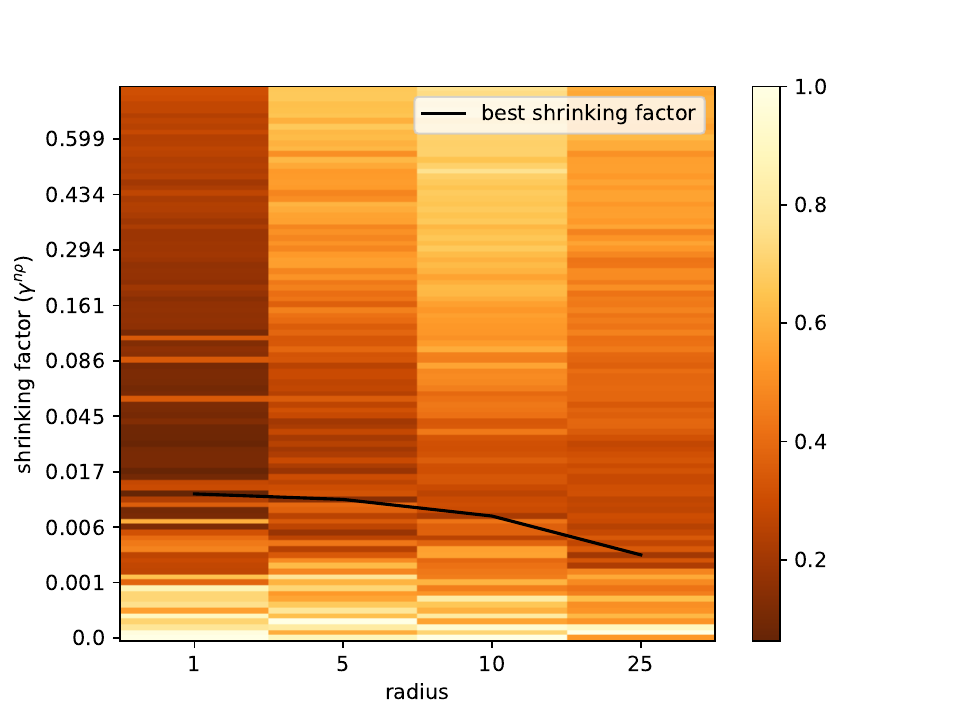}
        }
        \subfigure[Dimension $n=100$]{
            \includegraphics[width=0.22\linewidth]{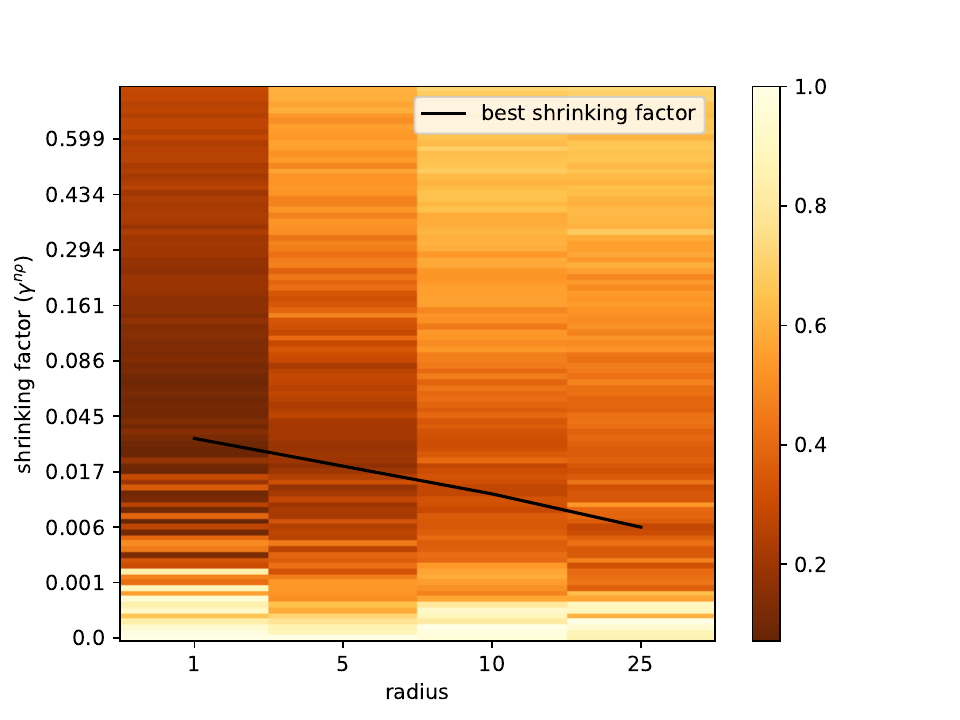}
        }
        \subfigure[Dimension $n=250$]{
            \includegraphics[width=0.22\linewidth]{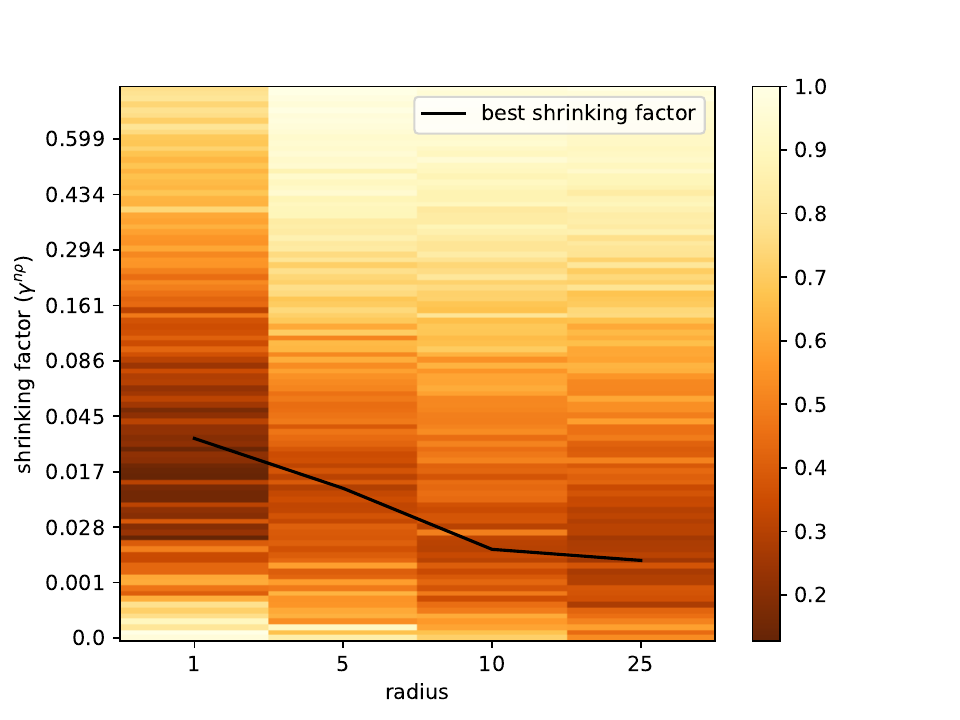}
        }
        \subfigure[Dimension $n=500$]{
            \includegraphics[width=0.22\linewidth]{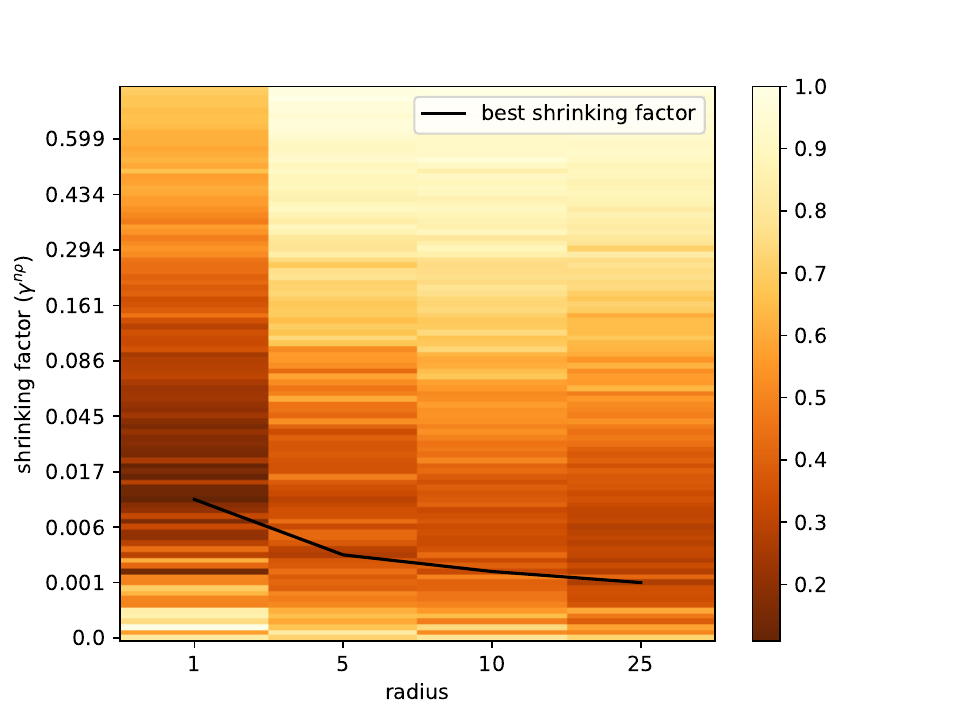}
        }
        \caption{\textbf{Comparison of shrinking factor $\mathbf{\gamma^{n\rho}}$ and radius $\mathbf{r}$ of the solution space $\mathbf{\Omega=[-r,r]^n}$. }{\small
        Results of different dimension are presented in each subfigure respectively. In each subfigure, the horizontal axis is the radius of solution space and the vertical axis is shrinking factor. Each pixel represents the heat of y-wise normalized mean function value at the $30n$ step. The black curve is the best shrinking factor of each solution space radius.
        }}
        \label{factorradius}
    \end{figure}
\end{itemize}

\begin{table}[htbp]
    \caption{\textbf{Comparison of shrinking frequencies $\mathbf{\rho}$ for Ackley on $\mathbf{\Omega=[-10,10]^n}$ with shrinking rate $\mathbf{\gamma=0.95}$.}\small{ Mean and standard deviation of function value at the $30n$ step are listed in the table. The first row of the table with $\rho=0$ is the results of \emph{RACOS} for reference. We omit the results of $\rho$ bigger than 0.1 for concision. The bold fonts are relative better results in each dimension.}}
    \resizebox{\linewidth}{.35\textheight}{
    % \tiny{
        \begin{tabular}{c|cccccccccc}
            \diagbox{$\mathbf{\rho}$}{$\mathbf{n}$}    & \textbf{50}      & \textbf{100}     & \textbf{150}     & \textbf{200}     & \textbf{250}     & \textbf{300}     & \textbf{350}     & \textbf{400}     & \textbf{450}     & \textbf{500}     \\ \hline
            \textbf{0}     & $ 3.8 \pm 0.2 $ & $ 3.9 \pm 0.2 $  & $ 5.9 \pm 0.1 $  & $ 5.7 \pm 0.1 $  & $ 5.8 \pm 0.2 $  & $ 5.8 \pm 0.1 $  & $ 5.9 \pm 0.0 $  & $ 5.8 \pm 0.0 $  & $ 5.9 \pm 0.1 $  & $ 5.8 \pm 0.1 $  \\
            \textbf{0.002} & $ 3.7 \pm 0.2 $ & $ 3.5 \pm 0.2 $ & $ 4.3 \pm 0.3 $ & $ 4.4 \pm 0.3 $ & $ 4.0 \pm 0.6 $ & $ 3.9 \pm 0.5 $  & $ 3.7 \pm 0.4 $  & $ 3.3 \pm 0.4 $  & $ 3.3 \pm 0.3 $  & $ 2.6 \pm 0.4 $  \\
            \textbf{0.004} & $ 3.4 \pm 0.3 $ & $ 3.2 \pm 0.1 $ & $ 3.8 \pm 0.6 $ & $ 3.7 \pm 0.2 $ & $ 2.8 \pm 0.5 $ & $ 2.1 \pm 0.5 $  & $ \mathbf{1.9 \pm 0.2} $  & $ \mathbf{1.9 \pm 0.2} $  & $ \mathbf{1.8 \pm 0.6} $  & $ \mathbf{1.7 \pm 0.4} $  \\
            \textbf{0.006} & $ 3.3 \pm 0.3 $ & $ 2.9 \pm 0.2 $ & $ 2.9 \pm 0.4 $ & $ 2.0 \pm 0.1 $ & $ 2.1 \pm 0.4 $ & $ \mathbf{1.8 \pm 0.2} $  & $ \mathbf{1.9 \pm 0.2} $  & $ 2.4 \pm 0.3 $  & $ 2.7 \pm 0.3 $  & $ 3.1 \pm 1.5 $  \\
            \textbf{0.008} & $ 3.0 \pm 0.3 $ & $ 2.4 \pm 0.4 $ & $ 2.3 \pm 0.5 $ & $ \mathbf{1.7 \pm 0.2} $ & $ \mathbf{1.8 \pm 0.4} $ & $ 2.3 \pm 0.7 $  & $ 2.5 \pm 0.4 $  & $ 3.9 \pm 0.8 $  & $ 4.5 \pm 0.9 $  & $ 5.5 \pm 1.0 $  \\
            \textbf{0.010} & $ 2.8 \pm 0.4 $ & $ \mathbf{1.8 \pm 0.5} $ & $ \mathbf{1.9 \pm 0.4} $ & $ 2.1 \pm 0.2 $ & $ 2.3 \pm 0.6 $ & $ 3.6 \pm 0.8 $  & $ 4.1 \pm 0.6 $  & $ 4.9 \pm 0.9 $  & $ 7.5 \pm 0.7 $  & $ 6.3 \pm 0.8 $  \\
            \textbf{0.012} & $ 2.5 \pm 0.1 $ & $ \mathbf{1.4 \pm 0.2} $ & $ \mathbf{1.6 \pm 0.4} $ & $ \mathbf{1.8 \pm 0.2} $ & $ 3.1 \pm 0.7 $ & $ 4.4 \pm 1.1 $  & $ 5.4 \pm 0.5 $  & $ 5.6 \pm 1.4 $  & $ 6.6 \pm 0.8 $  & $ 7.5 \pm 0.7 $  \\
            \textbf{0.014} & $ 2.6 \pm 0.3 $ & $ \mathbf{1.5 \pm 0.3} $ & $ 2.5 \pm 0.5 $ & $ 2.2 \pm 0.5 $ & $ 3.9 \pm 0.6 $ & $ 5.0 \pm 0.8 $  & $ 7.1 \pm 1.2 $  & $ 6.3 \pm 0.6 $  & $ 8.3 \pm 0.7 $  & $ 8.3 \pm 0.5 $  \\
            \textbf{0.016} & $ 2.6 \pm 0.4 $ & $ \mathbf{1.3 \pm 0.4} $ & $ 2.0 \pm 0.3 $ & $ 3.4 \pm 1.3 $ & $ 4.4 \pm 0.9 $ & $ 6.1 \pm 1.0 $  & $ 7.0 \pm 1.1 $  & $ 6.9 \pm 0.8 $  & $ 8.0 \pm 0.6 $  & $ 8.9 \pm 0.8 $  \\
            \textbf{0.018} & $ 2.3 \pm 0.2 $ & $ \mathbf{1.3 \pm 0.4} $ & $ 2.8 \pm 0.8 $ & $ 4.1 \pm 0.6 $ & $ 5.1 \pm 0.9 $ & $ 6.9 \pm 1.0 $  & $ 7.1 \pm 1.3 $  & $ 8.3 \pm 0.5 $  & $ 9.1 \pm 0.9 $  & $ 9.3 \pm 0.4 $  \\
            \textbf{0.020} & $ 2.0 \pm 0.6 $ & $ 2.0 \pm 0.5 $ & $ 3.2 \pm 1.2 $ & $ 4.4 \pm 0.8 $ & $ 6.3 \pm 1.3 $ & $ 7.2 \pm 1.1 $  & $ 7.4 \pm 0.6 $  & $ 9.1 \pm 0.8 $  & $ 10.2 \pm 0.9 $ & $ 10.0 \pm 0.6 $ \\
            \textbf{0.022} & $ \mathbf{1.7 \pm 0.3} $ & $ 2.0 \pm 1.0 $ & $ 3.9 \pm 1.3 $ & $ 5.3 \pm 1.0 $ & $ 6.8 \pm 1.0 $ & $ 7.5 \pm 1.1 $  & $ 8.8 \pm 0.6 $  & $ 9.1 \pm 0.8 $  & $ 10.7 \pm 0.5 $ & $ 10.0 \pm 0.3 $ \\
            \textbf{0.024} & $ \mathbf{1.9 \pm 0.2} $ & $ 3.3 \pm 0.9 $ & $ 4.3 \pm 1.1 $ & $ 6.0 \pm 0.9 $ & $ 7.0 \pm 0.3 $ & $ 8.5 \pm 0.7 $  & $ 9.3 \pm 0.4 $  & $ 10.1 \pm 0.7 $ & $ 10.7 \pm 0.5 $ & $ 10.8 \pm 0.6 $ \\
            \textbf{0.026} & $ \mathbf{1.5 \pm 0.4} $ & $ 2.7 \pm 1.2 $ & $ 4.3 \pm 0.6 $ & $ 7.0 \pm 0.7 $ & $ 8.3 \pm 0.4 $ & $ 9.0 \pm 0.4 $  & $ 9.8 \pm 0.7 $  & $ 10.1 \pm 0.7 $ & $ 10.7 \pm 0.6 $ & $ 11.6 \pm 0.4 $ \\
            \textbf{0.028} & $ \mathbf{1.3 \pm 0.2} $ & $ 3.8 \pm 0.6 $ & $ 4.9 \pm 0.7 $ & $ 7.2 \pm 0.7 $ & $ 8.8 \pm 1.0 $ & $ 8.8 \pm 0.9 $  & $ 9.5 \pm 0.5 $  & $ 10.7 \pm 0.4 $ & $ 10.9 \pm 0.2 $ & $ 11.3 \pm 0.4 $ \\
            \textbf{0.030} & $ \mathbf{1.3 \pm 0.4} $ & $ 4.0 \pm 0.4 $ & $ 5.0 \pm 0.4 $ & $ 7.1 \pm 0.9 $ & $ 8.2 \pm 1.0 $ & $ 9.2 \pm 0.6 $  & $ 10.3 \pm 0.5 $ & $ 10.6 \pm 1.0 $ & $ 10.9 \pm 0.4 $ & $ 11.8 \pm 0.4 $ \\
            \textbf{0.032} & $ \mathbf{1.5 \pm 0.5} $ & $ 6.1 \pm 1.2 $ & $ 6.2 \pm 1.1 $ & $ 7.3 \pm 0.6 $ & $ 9.1 \pm 0.8 $ & $ 9.9 \pm 0.5 $  & $ 10.6 \pm 0.4 $ & $ 10.5 \pm 0.8 $ & $ 11.2 \pm 0.6 $ & $ 12.0 \pm 0.3 $ \\
            \textbf{0.034} & $ \mathbf{1.9 \pm 0.6} $ & $ 5.1 \pm 0.8 $ & $ 5.9 \pm 0.7 $ & $ 8.2 \pm 0.7 $ & $ 9.0 \pm 0.3 $ & $ 10.4 \pm 0.4 $ & $ 10.6 \pm 0.6 $ & $ 11.2 \pm 0.4 $ & $ 11.9 \pm 0.5 $ & $ 12.1 \pm 0.4 $ \\
            \textbf{0.036} & $ \mathbf{1.5 \pm 0.2} $ & $ 6.4 \pm 0.5 $ & $ 6.8 \pm 0.6 $ & $ 7.9 \pm 1.0 $ & $ 9.3 \pm 1.0 $ & $ 10.4 \pm 0.4 $ & $ 10.9 \pm 0.3 $ & $ 11.3 \pm 0.5 $ & $ 11.9 \pm 0.3 $ & $ 12.0 \pm 0.3 $ \\
            \textbf{0.038} & $ 2.2 \pm 1.6 $ & $ 5.1 \pm 0.8 $ & $ 6.1 \pm 0.6 $ & $ 8.2 \pm 1.0 $ & $ 9.6 \pm 0.8 $ & $ 10.8 \pm 0.5 $ & $ 10.9 \pm 0.5 $ & $ 11.6 \pm 0.3 $ & $ 11.8 \pm 0.4 $ & $ 12.3 \pm 0.5 $ \\
            \textbf{0.040} & $ 2.2 \pm 1.1 $ & $ 7.3 \pm 1.6 $ & $ 7.6 \pm 0.4 $ & $ 8.7 \pm 0.4 $ & $ 9.3 \pm 0.7 $ & $ 10.4 \pm 0.5 $ & $ 11.4 \pm 0.4 $ & $ 11.7 \pm 0.6 $ & $ 12.3 \pm 0.2 $ & $ 12.2 \pm 0.3 $ \\
            \textbf{0.042} & $ 2.2 \pm 0.8 $ & $ 6.4 \pm 1.2 $  & $ 7.6 \pm 0.8 $  & $ 9.3 \pm 0.8 $  & $ 10.0 \pm 0.8 $ & $ 10.4 \pm 0.5 $ & $ 11.6 \pm 0.3 $ & $ 12.3 \pm 0.4 $ & $ 12.0 \pm 0.5 $ & $ 12.5 \pm 0.5 $ \\
            \textbf{0.044} & $ \mathbf{1.8 \pm 0.6} $ & $ 6.8 \pm 1.0 $  & $ 6.9 \pm 0.9 $  & $ 9.6 \pm 0.5 $  & $ 10.3 \pm 0.4 $ & $ 11.0 \pm 1.0 $ & $ 11.1 \pm 0.3 $ & $ 12.0 \pm 0.3 $ & $ 12.4 \pm 0.4 $ & $ 12.6 \pm 0.3 $ \\
            \textbf{0.046} & $ 2.1 \pm 0.3 $ & $ 7.4 \pm 0.7 $  & $ 8.1 \pm 0.8 $  & $ 9.4 \pm 0.3 $  & $ 10.4 \pm 0.9 $ & $ 11.3 \pm 0.4 $ & $ 11.6 \pm 0.5 $ & $ 12.1 \pm 0.4 $ & $ 12.4 \pm 0.2 $ & $ 12.7 \pm 0.2 $ \\
            \textbf{0.048} & $ 3.2 \pm 1.1 $ & $ 6.6 \pm 1.0 $  & $ 7.5 \pm 1.3 $  & $ 9.9 \pm 0.4 $  & $ 10.1 \pm 0.4 $ & $ 11.3 \pm 0.2 $ & $ 12.2 \pm 0.4 $ & $ 12.0 \pm 0.5 $ & $ 12.3 \pm 0.3 $ & $ 12.5 \pm 0.3 $ \\
            \textbf{0.050} & $ 3.5 \pm 1.0 $ & $ 7.7 \pm 0.8 $  & $ 8.6 \pm 0.3 $  & $ 10.0 \pm 0.3 $ & $ 10.6 \pm 0.6 $ & $ 11.1 \pm 0.5 $ & $ 12.4 \pm 0.2 $ & $ 12.5 \pm 0.4 $ & $ 12.6 \pm 0.3 $ & $ 12.8 \pm 0.2 $ \\
            \textbf{0.052} & $ 3.6 \pm 0.7 $ & $ 8.8 \pm 0.9 $  & $ 8.1 \pm 0.6 $  & $ 10.0 \pm 0.8 $ & $ 11.0 \pm 0.4 $ & $ 11.6 \pm 0.7 $ & $ 12.7 \pm 0.2 $ & $ 12.5 \pm 0.1 $ & $ 12.8 \pm 0.4 $ & $ 13.3 \pm 0.1 $ \\
            \textbf{0.054} & $ 3.8 \pm 1.5 $ & $ 7.3 \pm 0.9 $  & $ 8.5 \pm 0.3 $  & $ 10.1 \pm 0.9 $ & $ 11.3 \pm 0.3 $ & $ 11.7 \pm 0.2 $ & $ 12.7 \pm 0.3 $ & $ 12.9 \pm 0.2 $ & $ 12.8 \pm 0.2 $ & $ 13.0 \pm 0.3 $ \\
            \textbf{0.056} & $ 3.8 \pm 1.3 $ & $ 8.8 \pm 1.0 $  & $ 9.1 \pm 0.8 $  & $ 10.5 \pm 0.7 $ & $ 11.1 \pm 0.7 $ & $ 11.8 \pm 0.4 $ & $ 12.4 \pm 0.2 $ & $ 12.6 \pm 0.4 $ & $ 13.1 \pm 0.1 $ & $ 13.0 \pm 0.2 $ \\
            \textbf{0.058} & $ 4.1 \pm 1.0 $ & $ 9.1 \pm 1.1 $  & $ 9.3 \pm 1.3 $  & $ 10.7 \pm 0.5 $ & $ 10.9 \pm 0.2 $ & $ 11.9 \pm 0.3 $ & $ 12.1 \pm 0.4 $ & $ 12.7 \pm 0.4 $ & $ 13.1 \pm 0.2 $ & $ 13.2 \pm 0.3 $ \\
            \textbf{0.060} & $ 4.1 \pm 1.0 $ & $ 8.8 \pm 0.4 $  & $ 9.1 \pm 0.9 $  & $ 10.8 \pm 0.5 $ & $ 11.2 \pm 0.5 $ & $ 11.8 \pm 0.3 $ & $ 12.5 \pm 0.2 $ & $ 12.8 \pm 0.2 $ & $ 13.0 \pm 0.3 $ & $ 13.2 \pm 0.2 $ \\
            \textbf{0.062} & $ 4.1 \pm 1.7 $ & $ 9.2 \pm 1.0 $  & $ 9.1 \pm 0.6 $  & $ 10.9 \pm 0.5 $ & $ 11.9 \pm 0.5 $ & $ 12.1 \pm 0.3 $ & $ 12.4 \pm 0.2 $ & $ 12.9 \pm 0.4 $ & $ 13.0 \pm 0.3 $ & $ 13.3 \pm 0.1 $ \\
            \textbf{0.064} & $ 4.5 \pm 1.2 $ & $ 8.6 \pm 0.5 $  & $ 9.7 \pm 0.4 $  & $ 11.1 \pm 0.8 $ & $ 11.7 \pm 0.2 $ & $ 12.3 \pm 0.5 $ & $ 12.6 \pm 0.3 $ & $ 13.1 \pm 0.2 $ & $ 13.3 \pm 0.2 $ & $ 13.4 \pm 0.2 $ \\
            \textbf{0.066} & $ 4.7 \pm 0.3 $ & $ 9.5 \pm 0.9 $  & $ 9.2 \pm 0.4 $  & $ 11.0 \pm 0.5 $ & $ 12.0 \pm 0.3 $ & $ 12.1 \pm 0.3 $ & $ 12.8 \pm 0.4 $ & $ 12.9 \pm 0.2 $ & $ 13.3 \pm 0.2 $ & $ 13.3 \pm 0.2 $ \\
            \textbf{0.068} & $ 4.7 \pm 0.7 $ & $ 9.2 \pm 1.0 $  & $ 9.7 \pm 0.7 $  & $ 11.0 \pm 0.7 $ & $ 11.7 \pm 0.4 $ & $ 12.8 \pm 0.4 $ & $ 12.5 \pm 0.6 $ & $ 13.0 \pm 0.2 $ & $ 13.4 \pm 0.1 $ & $ 13.4 \pm 0.2 $ \\
            \textbf{0.070} & $ 5.4 \pm 1.5 $ & $ 9.5 \pm 0.9 $  & $ 10.1 \pm 0.6 $ & $ 10.8 \pm 0.4 $ & $ 12.3 \pm 0.3 $ & $ 12.4 \pm 0.4 $ & $ 12.5 \pm 0.7 $ & $ 13.1 \pm 0.4 $ & $ 13.3 \pm 0.2 $ & $ 13.5 \pm 0.1 $ \\
            \textbf{0.072} & $ 5.3 \pm 1.2 $ & $ 9.1 \pm 0.8 $  & $ 10.1 \pm 0.4 $ & $ 11.7 \pm 0.5 $ & $ 12.2 \pm 0.6 $ & $ 12.5 \pm 0.4 $ & $ 13.0 \pm 0.4 $ & $ 13.3 \pm 0.2 $ & $ 13.2 \pm 0.2 $ & $ 13.7 \pm 0.2 $ \\
            \textbf{0.074} & $ 5.8 \pm 1.0 $ & $ 10.1 \pm 0.8 $ & $ 10.3 \pm 0.4 $ & $ 11.4 \pm 0.3 $ & $ 12.0 \pm 0.4 $ & $ 12.5 \pm 0.4 $ & $ 12.7 \pm 0.3 $ & $ 13.0 \pm 0.4 $ & $ 13.3 \pm 0.3 $ & $ 13.7 \pm 0.1 $ \\
            \textbf{0.076} & $ 5.2 \pm 1.3 $ & $ 9.8 \pm 0.5 $  & $ 10.6 \pm 0.5 $ & $ 11.3 \pm 0.9 $ & $ 12.3 \pm 0.3 $ & $ 12.7 \pm 0.5 $ & $ 13.0 \pm 0.2 $ & $ 13.2 \pm 0.2 $ & $ 13.5 \pm 0.3 $ & $ 13.7 \pm 0.2 $ \\
            \textbf{0.078} & $ 5.9 \pm 0.5 $ & $ 10.3 \pm 0.5 $ & $ 9.8 \pm 0.6 $  & $ 11.8 \pm 0.1 $ & $ 12.1 \pm 0.1 $ & $ 12.7 \pm 0.4 $ & $ 13.2 \pm 0.3 $ & $ 13.3 \pm 0.3 $ & $ 13.6 \pm 0.1 $ & $ 13.7 \pm 0.2 $ \\
            \textbf{0.080} & $ 5.6 \pm 0.7 $ & $ 10.1 \pm 0.1 $ & $ 10.6 \pm 0.4 $ & $ 11.4 \pm 0.4 $ & $ 12.3 \pm 0.5 $ & $ 13.0 \pm 0.3 $ & $ 13.1 \pm 0.1 $ & $ 13.4 \pm 0.3 $ & $ 13.5 \pm 0.4 $ & $ 13.7 \pm 0.2 $ \\
            \textbf{0.082} & $ 4.3 \pm 1.3 $ & $ 10.3 \pm 0.8 $ & $ 10.2 \pm 0.7 $ & $ 11.8 \pm 0.5 $ & $ 12.5 \pm 0.4 $ & $ 12.8 \pm 0.3 $ & $ 13.1 \pm 0.3 $ & $ 13.4 \pm 0.2 $ & $ 13.5 \pm 0.3 $ & $ 13.8 \pm 0.2 $ \\
            \textbf{0.084} & $ 6.7 \pm 0.9 $ & $ 10.6 \pm 0.3 $ & $ 10.9 \pm 0.2 $ & $ 11.8 \pm 0.3 $ & $ 12.5 \pm 0.3 $ & $ 13.0 \pm 0.3 $ & $ 13.3 \pm 0.2 $ & $ 13.4 \pm 0.3 $ & $ 13.6 \pm 0.2 $ & $ 13.8 \pm 0.2 $ \\
            \textbf{0.086} & $ 4.9 \pm 0.6 $ & $ 10.2 \pm 0.6 $ & $ 11.0 \pm 0.4 $ & $ 11.9 \pm 0.3 $ & $ 12.4 \pm 0.2 $ & $ 12.6 \pm 0.5 $ & $ 13.0 \pm 0.3 $ & $ 13.5 \pm 0.2 $ & $ 13.7 \pm 0.2 $ & $ 13.9 \pm 0.2 $ \\
            \textbf{0.088} & $ 5.8 \pm 1.0 $ & $ 10.7 \pm 0.6 $ & $ 10.9 \pm 0.2 $ & $ 11.7 \pm 0.6 $ & $ 12.5 \pm 0.2 $ & $ 13.0 \pm 0.5 $ & $ 13.3 \pm 0.2 $ & $ 13.6 \pm 0.3 $ & $ 13.6 \pm 0.1 $ & $ 13.9 \pm 0.1 $ \\
            \textbf{0.090} & $ 6.6 \pm 1.4 $ & $ 10.2 \pm 0.6 $ & $ 11.1 \pm 0.4 $ & $ 12.1 \pm 0.3 $ & $ 12.6 \pm 0.5 $ & $ 13.0 \pm 0.2 $ & $ 13.5 \pm 0.1 $ & $ 13.4 \pm 0.2 $ & $ 13.6 \pm 0.2 $ & $ 13.8 \pm 0.2 $ \\
            \textbf{0.092} & $ 7.0 \pm 1.0 $ & $ 10.4 \pm 0.6 $ & $ 11.1 \pm 0.7 $ & $ 12.2 \pm 0.3 $ & $ 12.8 \pm 0.3 $ & $ 13.0 \pm 0.2 $ & $ 13.3 \pm 0.2 $ & $ 13.5 \pm 0.3 $ & $ 13.7 \pm 0.2 $ & $ 13.8 \pm 0.2 $ \\
            \textbf{0.094} & $ 7.9 \pm 0.5 $ & $ 10.2 \pm 0.2 $ & $ 11.2 \pm 0.6 $ & $ 12.3 \pm 0.2 $ & $ 12.5 \pm 0.3 $ & $ 12.8 \pm 0.3 $ & $ 13.2 \pm 0.2 $ & $ 13.5 \pm 0.2 $ & $ 13.6 \pm 0.2 $ & $ 13.9 \pm 0.2 $ \\
            \textbf{0.096} & $ 6.7 \pm 0.5 $ & $ 10.9 \pm 0.6 $ & $ 11.1 \pm 0.2 $ & $ 12.2 \pm 0.5 $ & $ 12.8 \pm 0.2 $ & $ 13.1 \pm 0.3 $ & $ 13.4 \pm 0.3 $ & $ 13.4 \pm 0.2 $ & $ 13.8 \pm 0.3 $ & $ 13.9 \pm 0.1 $ \\
            \textbf{0.098} & $ 7.6 \pm 0.5 $ & $ 10.7 \pm 0.5 $ & $ 11.1 \pm 0.2 $ & $ 12.2 \pm 0.3 $ & $ 12.6 \pm 0.3 $ & $ 13.0 \pm 0.4 $ & $ 13.3 \pm 0.2 $ & $ 13.6 \pm 0.2 $ & $ 13.7 \pm 0.2 $ & $ 14.0 \pm 0.1 $ \\
            \textbf{0.100}  & $ 8.2 \pm 1.0 $ & $ 10.8 \pm 0.3 $ & $ 11.3 \pm 0.8 $ & $ 11.9 \pm 0.4 $ & $ 13.0 \pm 0.2 $ & $ 13.3 \pm 0.3 $ & $ 13.4 \pm 0.1 $ & $ 13.6 \pm 0.2 $ & $ 13.8 \pm 0.1 $ & $ 13.8 \pm 0.2 $
        \end{tabular}}\label{table::1}
    % }
\end{table}

\newpage

% \section{Black-Box Tuning for LMaaS} \label{appendixLLM}
\begin{figure}[htb]
    \centering
    \subfigure[Yelp P.]{
        \includegraphics[width=0.3\linewidth]{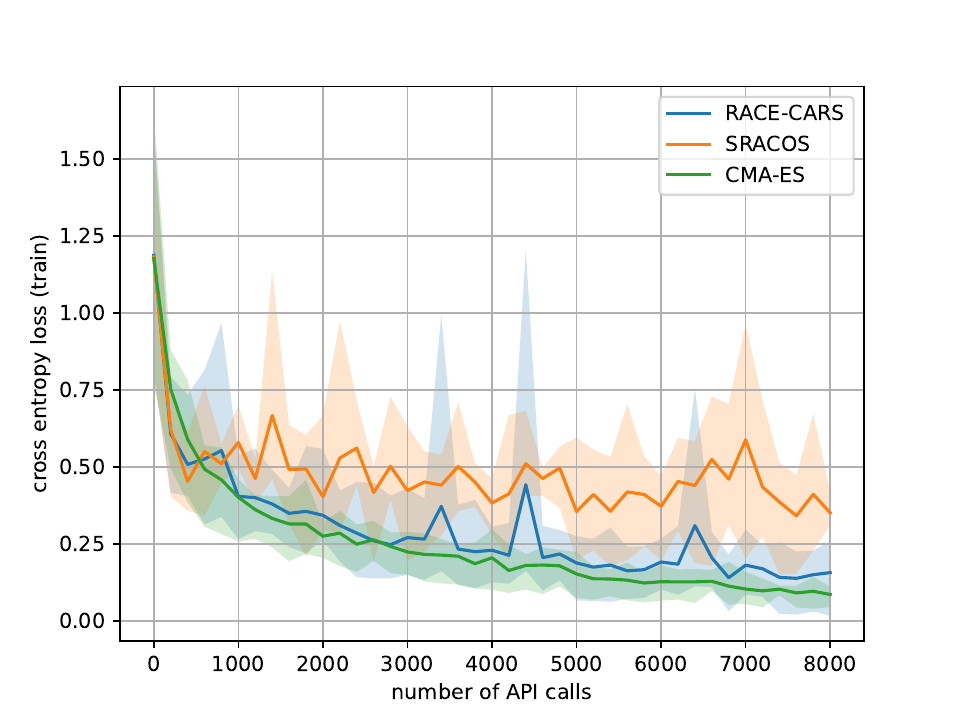}
    }
    \subfigure[AG's News]{
        \includegraphics[width=0.3\linewidth]{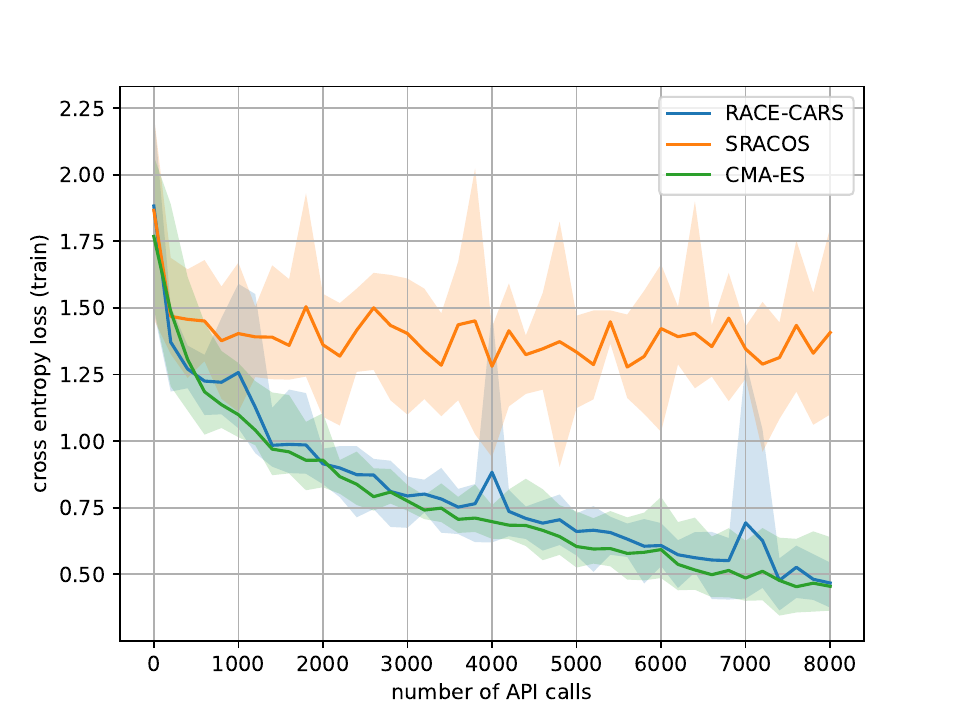}
    }
    \subfigure[RTE]{
        \includegraphics[width=0.3\linewidth]{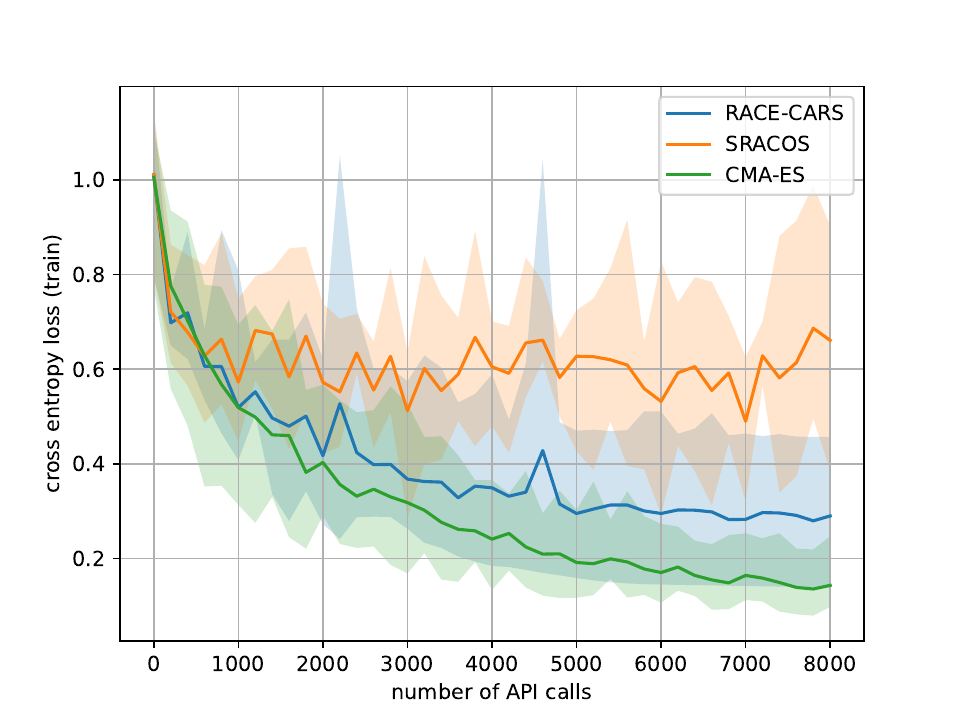}
    }\\
   \subfigure[Yelp P.]{
       \includegraphics[width=0.3\linewidth]{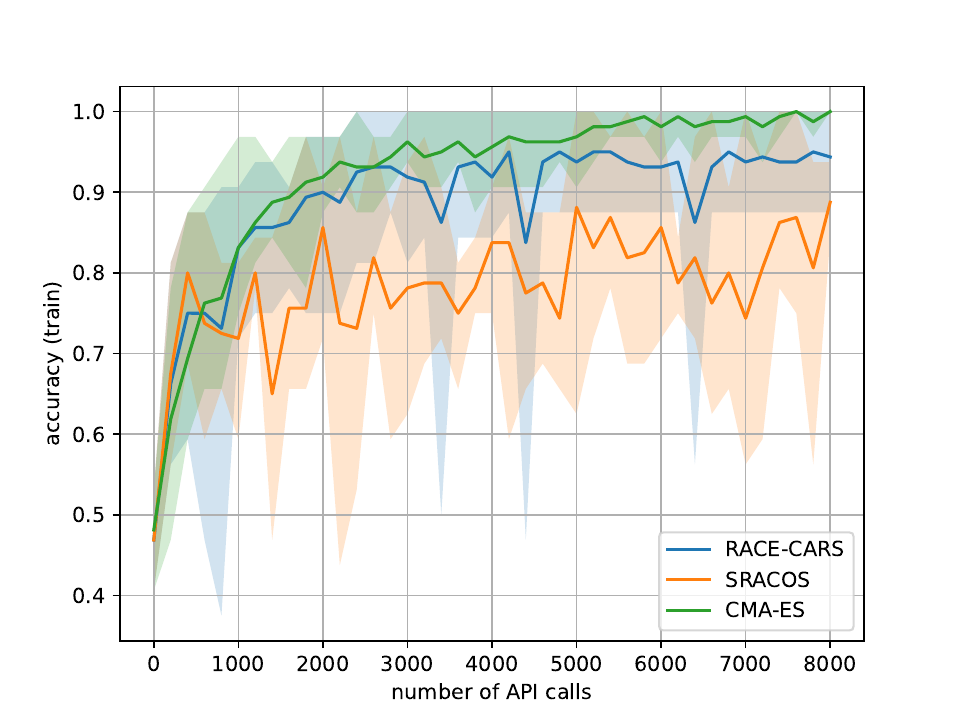}
   }
   \subfigure[AG's News]{
       \includegraphics[width=0.3\linewidth]{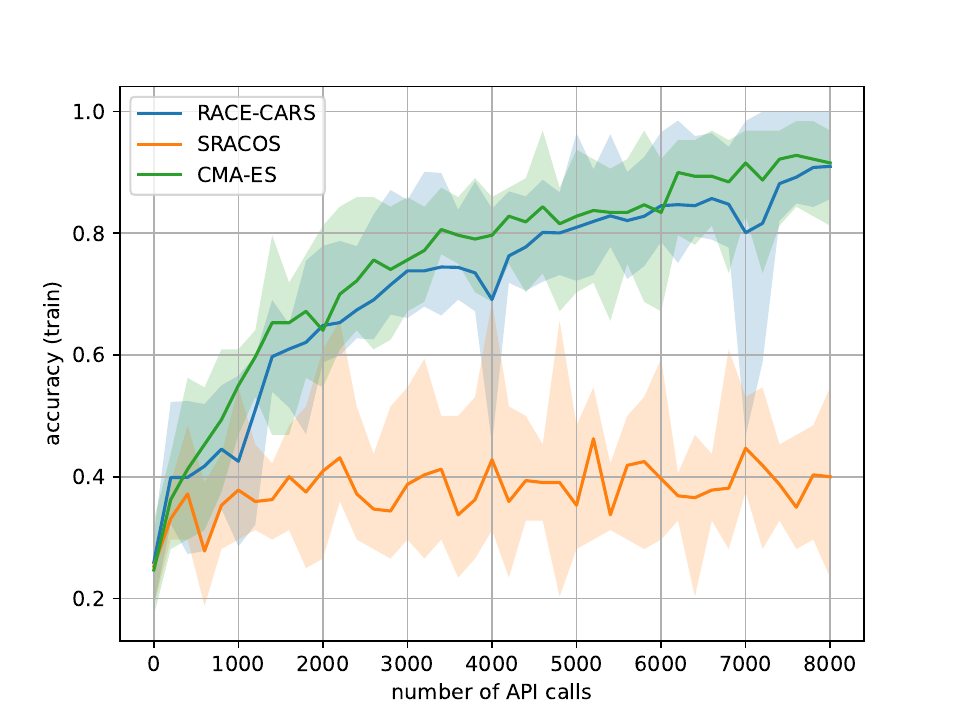}
   }
   \subfigure[RTE]{
       \includegraphics[width=0.3\linewidth]{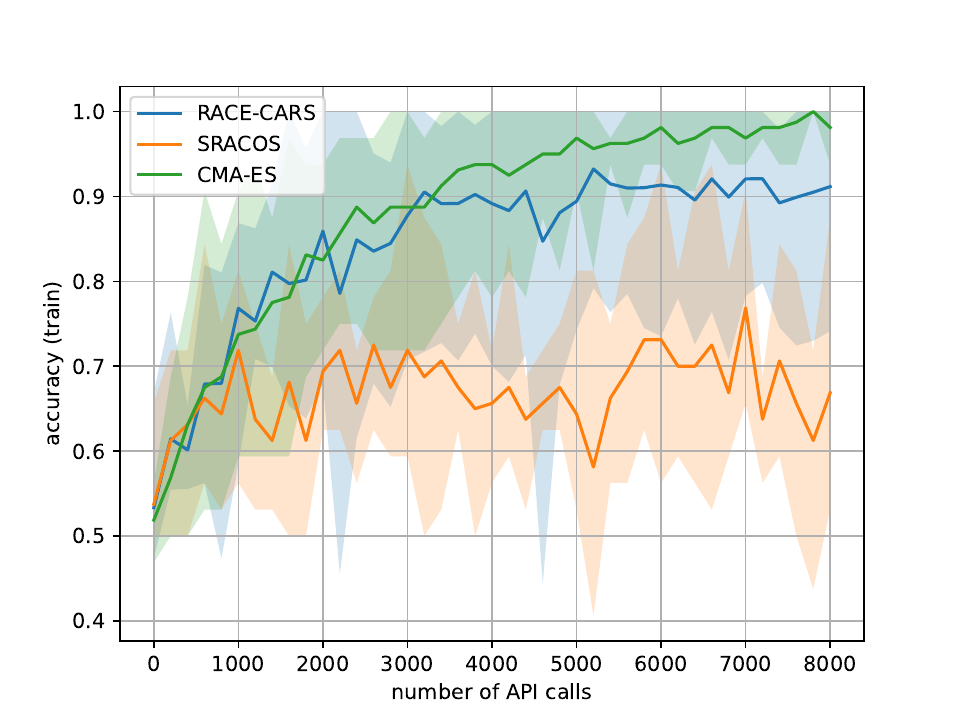}
   }\\
   \subfigure[Yelp P.]{
       \includegraphics[width=0.3\linewidth]{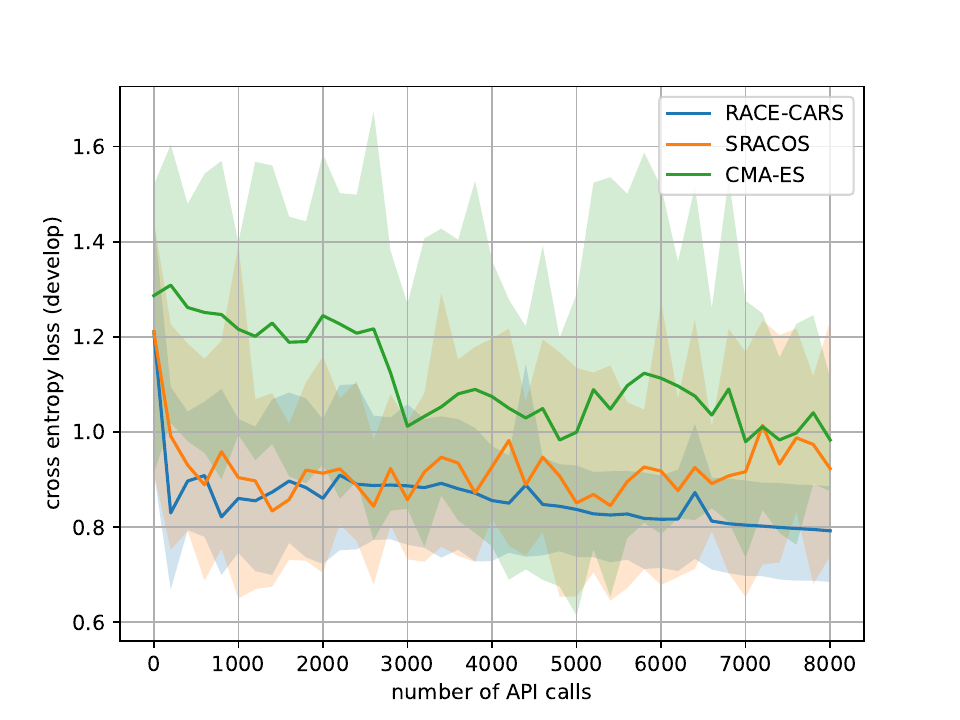}
   }
   \subfigure[AG's News]{
       \includegraphics[width=0.3\linewidth]{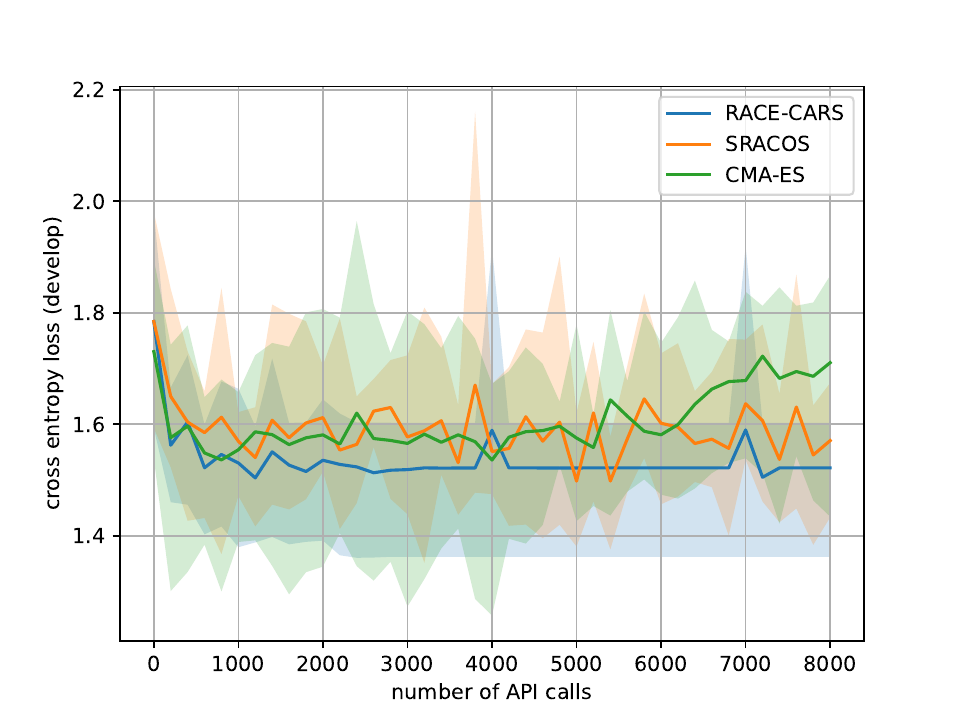}
   }
   \subfigure[RTE]{
       \includegraphics[width=0.3\linewidth]{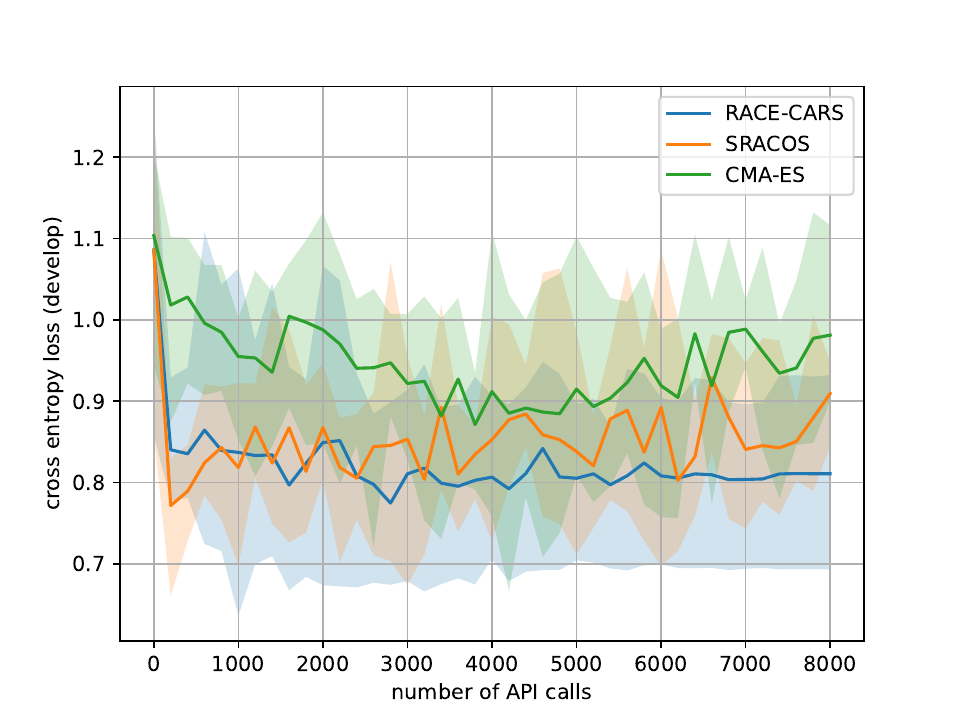}
   }\\
   \subfigure[Yelp P.]{
       \includegraphics[width=0.3\linewidth]{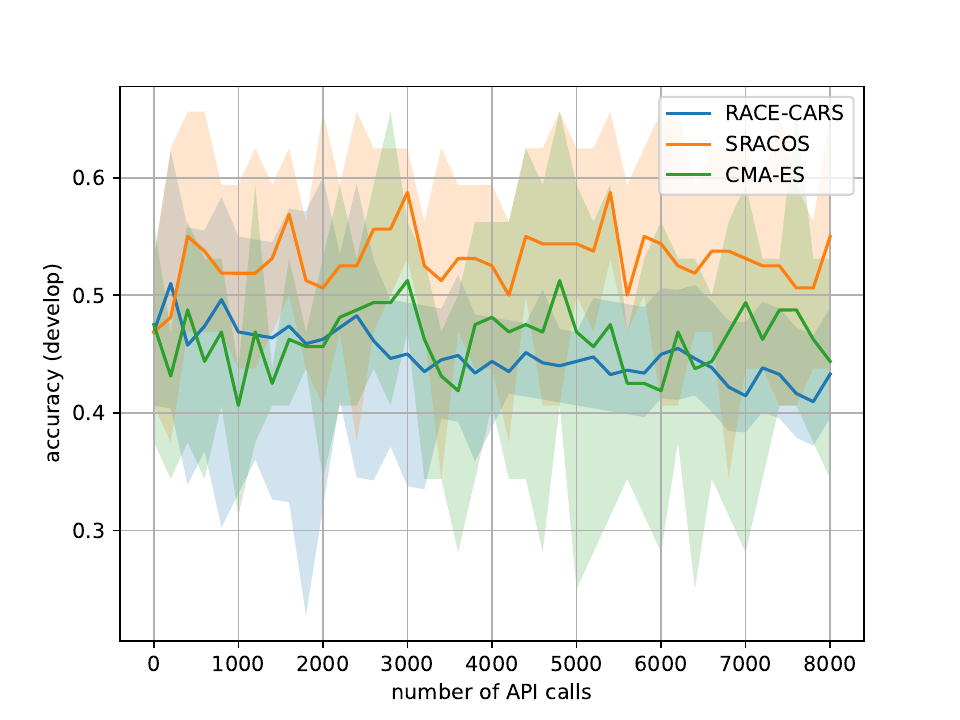}
   }
   \subfigure[AG's News]{
       \includegraphics[width=0.3\linewidth]{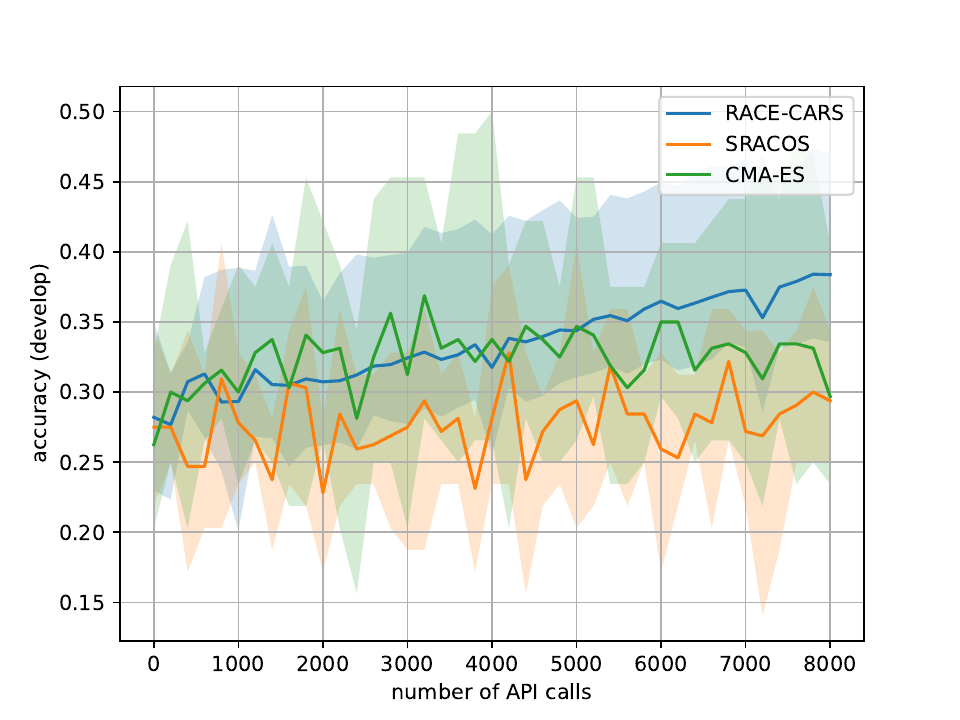}
   }
   \subfigure[RTE]{
       \includegraphics[width=0.3\linewidth]{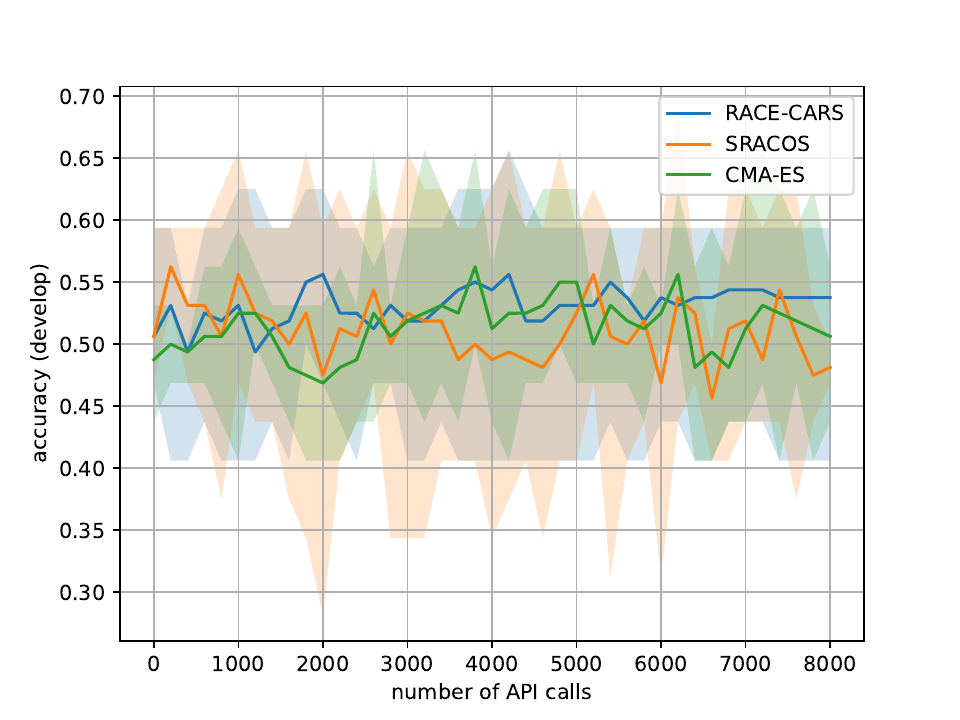}
   }
   \caption{\textbf{Black-Box Tuning for LMaaS.} {\small
   Results of Training Loss, Training Accuracy, Development Loss and Development Accuracy on Yelp P, AG's News and RTE.
   }}
   \label{developacc}
\end{figure} 

\newpage
\thispagestyle{empty}
\clearpage
\section{Theory Supplementary}
\subsection{Proofs of Theorems}

\addtocounter{theorem}{-2}
   
\begin{theorem}\label{appendixsracosconvergence}{\rm
    For sequential-mode classification-based DFO \Algref{sequentialclassification}, let $\rmX_t=\rmX_{h_t},$ $\epsilon>0$ and $0<\delta<1.$ When $\Omega_\epsilon$ is $\eta$-shattered by $h_t$ for all $t=r+1\ldots,T$ and $\max\limits_{t=r+1,\ldots,T}\sP(\{x\in\Omega\colon h_t(x)=1\})\le p\le 1,$ the $(\epsilon,\delta)$-query complexity is upper bounded by 
    \begin{equation*}
        \mathcal{O}\bigg(\max\{\bigl(\lambda\frac{\eta}{p}+(1-\lambda)\bigr)^{-1}\bigl(\frac{1}{|\Omega_\epsilon|}\ln\frac{1}{\delta}-r\bigr)+r,T\}\bigg).
    \end{equation*}
}
\end{theorem}
   
\begin{proof}
    Let $\tilde{x}:=\arg\min\limits_{t=1,\ldots,T}~f(x_t),$ then
    \begin{align*}
        &\Pr\left(f(\tilde{x})-f^*>\epsilon\right)\\
        =&\E\left[\sI_{\{\rmY_1,\ldots,\rmY_{T-1}\in\Omega_\epsilon^c\}}\E[\sI_{\{\rmY_T\in\Omega_\epsilon^c\}}|\gF_{T-1}]\right]\\
        =&\E\bigl[\sI_{\{\rmY_1,\ldots,\rmY_{T-2}\in\Omega_\epsilon^c\}}\E\left[\sI_{\{\rmY_{T-1}\in\Omega_\epsilon^c\}}\E[\sI_{\{\rmY_T\in\Omega_\epsilon^c\}}|\gF_{T-1}]|\gF_{T-2}\right]\bigr]\\
        =&\cdots\cdots\\
        =&\E\bigl[\sI_{\{\rmY_1,\ldots,\rmY_r\in\Omega_\epsilon^c\}}\E\big[\sI_{\{\rmY_{r+1}\in\Omega_\epsilon^c\}}\cdots\E\left[\sI_{\{\rmY_{T-1}\in\Omega_\epsilon^c\}}\E[\sI_{\{\rmY_T\in\Omega_\epsilon^c\}}|\gF_{T-1}]|\gF_{T-2}\right]\cdots|\gF_{r}\big]\bigr].
    \end{align*}
    Where $\sI_B(x)$ is the identical function on $B\in\gF$ such that $\sI_B(x)\equiv 1$ for all $x\in B$ and $\sI_B(x)\equiv0$ otherwise. At step $t\ge r+1,$ since $\rmX_\Omega$ is independent to $\gF_{t-1},$ it holds
    \begin{align*}
        \E[\I_{\{\Y_t\in\Omega_\epsilon^c\}}|\F_{t-1}]
        =&\E[\I_{\{\lambda \X_t+(1-\lambda)\X_\Omega\in\Omega_\epsilon^c\}}|\F_{t-1}]\\
        =&\lambda(1-\E[\I_{\{\X_{h_t}\in\Omega_\epsilon\}}|\F_{t-1}])+(1-\lambda)(1-|\Omega_\epsilon|).
    \end{align*}
    Under the assumption that $\Omega_\epsilon$ is $\eta$-shattered by $h_t,$ it holds the relation that
    \begin{equation*}
        \E[\I_{\{\X_{h_t}\in\Omega_\epsilon\}}|\F_{t-1}]=\frac{\p(\{x\in\Omega_\epsilon\colon h_t(x)=1\})}{\p(\{x\in\Omega\colon h_t(x)=1\})}\ge\frac{\eta}{p}|\Omega_\epsilon|.
    \end{equation*}
    Therefore,
    \begin{align*}
        \E[\I_{\{\Y_t\in\Omega_\epsilon^c\}}|\F_{t-1}]
        =&\lambda(1-\E[\I_{\{\X_{h_t}\in\Omega_\epsilon\}}|\F_{t-1}])+(1-\lambda)(1-|\Omega_\epsilon|)\\
        \le&1-\bigl(\lambda\frac{\eta}{p}+(1-\lambda)|\bigr)\Omega_\epsilon|.
    \end{align*}
    Apparently, the upper bound of $\E[\I_{\{\Y_t\in\Omega_\epsilon^c\}}|\F_{t-1}]$ satisfies $0<1-\bigl(\lambda\frac{\eta}{p}+(1-\lambda)\bigr)|\Omega_\epsilon|<1,$ thus
    \begin{align*}
       \E\big[\I_{\{\Y_t\in\Omega_\epsilon^c\}}\E[\I_{\{\Y_{t+1}\in\Omega_\epsilon^c\}}|\F_t]|\F_{t-1}\big]
       \le&\bigl(1-(\lambda\frac{\eta}{p}+(1-\lambda))|\Omega_\epsilon|\bigr)\E[\I_{\{Y_t\in\Omega_\epsilon^c\}}|\F_{t-1}]\\
       \le&\bigl(1-(\lambda\frac{\eta}{p}+(1-\lambda))|\Omega_\epsilon|\bigr)^2.
   \end{align*}
   Moreover,
   \begin{align*}
       &\Pr\left(f(\tilde{x})-f^*>\epsilon\right)\\
       =&\E\bigl[\I_{\{\Y_1,\ldots,\Y_r\in\Omega_\epsilon^c\}}\E\big[\I_{\{\Y_{r+1}\in\Omega_\epsilon^c\}}\cdots\E\left[\I_{\{\Y_{T-1}\in\Omega_\epsilon^c\}}\E[\I_{\{\Y_T\in\Omega_\epsilon^c\}}|\F_{T-1}]|\F_{T-2}\right]\cdots|\F_{r}\big]\bigr]\\
       \le&\bigl(1-(\lambda\frac{\eta}{p}+(1-\lambda))|\Omega_\epsilon|\bigr)^{T-r}\E[\Y_1,\ldots,\Y_r\in\Omega_\epsilon^c]\\
       =&\bigl(1-(\lambda\frac{\eta}{p}+(1-\lambda))|\Omega_\epsilon|\bigr)^{T-r}(1-|\Omega_\epsilon|)^r\\
       \le&\exp\left\{-\left((T-r)(\lambda\frac{\eta}{p}+(1-\lambda))+r\right)|\Omega_\epsilon|\right\}.
   \end{align*}
   In order that $\Pr\left(f(\tilde{x})-f^*>\epsilon\right)\le\delta,$ it suffices that
   \begin{equation*}
       \exp\left\{-\left((T-r)(\lambda\frac{\eta}{p}+(1-\lambda))+r\right)|\Omega_\epsilon|\right\}\le\delta,
   \end{equation*}
   hence the $(\epsilon,\delta)$-query complexity is upper bounded by 
   \begin{equation*}
       \mathcal{O}\bigg(\max\{\bigl(\lambda\frac{\eta}{p}+(1-\lambda)\bigr)^{-1}\bigl(\frac{1}{|\Omega_\epsilon|}\ln\frac{1}{\delta}-r\bigr)+r,T\}\bigg).
   \end{equation*}
\end{proof}

\begin{theorem}\label{appendixracecarsconvergence}{\rm
    For \Algref{racecars} with region shrinking rate $0<\gamma<1$ and region shrinking frequency $0<\rho<1.$ Let $\epsilon>0$ and $0<\delta<1.$ When $\Omega_\epsilon$ is $\eta$-shattered by $\tilde{h}_t$ for all $t=r+1\ldots,T,$ the $(\epsilon,\delta)$-query complexity is upper bounded by 
    \begin{equation*}
        \mathcal{O}\bigg(\max\{\bigl(\frac{\gamma^{-\rho}+\gamma^{-(T-r)\rho}}{2}\lambda\eta+(1-\lambda)\bigr)^{-1}\bigl(\frac{1}{|\Omega_\epsilon|}\ln\frac{1}{\delta}-r\bigr)+r,T\}\bigg).
        \end{equation*}
    }
\end{theorem}
   
\begin{proof}
    Let $\tilde{x}:=\arg\min\limits_{t=1,\ldots,T}~f(x_t),$ then
    \begin{align*}
        &\Pr\left(f(\tilde{x})-f^*>\epsilon\right)\\
        =&\E\left[\I_{\{\Y_1,\ldots,\Y_{T-1}\in\Omega_\epsilon^c\}}\E[\I_{\{\Y_T\in\Omega_\epsilon^c\}}|\F_{T-1}]\right]\\
        =&\E\bigl[\I_{\{\Y_1,\ldots,Y_{T-2}\in\Omega_\epsilon^c\}}\E\left[\I_{\{\Y_{T-1}\in\Omega_\epsilon^c\}}\E[\I_{\{\Y_T\in\Omega_\epsilon^c\}}|\F_{T-1}]|\F_{T-2}\right]\bigr]\\
        =&\cdots\cdots\\
        =&\E\bigl[\I_{\{\Y_1,\ldots,\Y_r\in\Omega_\epsilon^c\}}\E\big[\I_{\{\Y_{r+1}\in\Omega_\epsilon^c\}}\cdots\E\left[\I_{\{\Y_{T-1}\in\Omega_\epsilon^c\}}\E[\I_{\{\Y_T\in\Omega_\epsilon^c\}}|\F_{T-1}]|\F_{T-2}\right]\cdots|\F_{r}\big]\bigr].
    \end{align*}
    At step $t\ge r+1,$ since $\X_\Omega$ is independent to $\F_{t-1},$ it holds
    \begin{align*}
        \E[\I_{\{\Y_t\in\Omega_\epsilon^c\}}|\F_{t-1}]
        =&\E[\I_{\{\lambda \X_t+(1-\lambda)\X_\Omega\in\Omega_\epsilon^c\}}|\F_{t-1}]\\
        =&\lambda(1-\E[\I_{\{\X_t\in\Omega_\epsilon\}}|\F_{t-1}])+(1-\lambda)(1-|\Omega_\epsilon|).
    \end{align*}
    The expectation of probability that $\tilde{h}_t$ hits active region is upper bounded by
    \begin{equation*}
        \E\bigl[\p(\{x\in\Omega\colon\tilde{h}_t(x)=1\})|\F_{t-1}\bigr]\le\gamma^{(t-r)\rho}\p[\Omega]=\gamma^{(t-r)\rho}.
    \end{equation*}
    Under the assumption that $\Omega_\epsilon$ is $\eta$-shattered by $\tilde{h}_t,$ it holds the relation that
    \begin{align*}
        \E\big[\I_{\{\X_t\in\Omega_\epsilon\}}|\F_{t-1}\big]=&\frac{\p\bigl(\{x\in\Omega_\epsilon\colon\tilde{h}_t(x)=1\}\bigr)}{\E\bigl[\p(\{x\in\Omega\colon\tilde{h}_t(x)=1\})|\F_{t-1}\bigr]}\\
        \ge&\gamma^{-(t-r)\rho}\eta|\Omega_\epsilon|.
    \end{align*}
    Therefore,
    \begin{equation*}
        \E[\I_{\{\Y_t\in\Omega_\epsilon^c\}}|\F_{t-1}]\le1-\bigl(\lambda\gamma^{-(t-r)\rho}\eta+(1-\lambda)|\bigr)\Omega_\epsilon|.
    \end{equation*}
    Moreover,
    \begin{align*}
        &\Pr\left(f(\tilde{x})-f^*>\epsilon\right)\\
        =&\E\bigl[\I_{\{\Y_1,\ldots,\Y_r\in\Omega_\epsilon^c\}}\E\big[\I_{\{\Y_{r+1}\in\Omega_\epsilon^c\}}\cdots\E\left[\I_{\{\Y_{T-1}\in\Omega_\epsilon^c\}}\E[\I_{\{\Y_T\in\Omega_\epsilon^c\}}|\F_{T-1}]|\F_{T-2}\right]\cdots|\F_{r}\big]\bigr]\\
        \le&\prod_{t=r+1}^{T}\bigl(1-\bigl(\lambda\gamma^{-(t-r)\rho}\eta+(1-\lambda)|\bigr)\Omega_\epsilon|\bigr)(1-|\Omega_\epsilon|)^r\\
        \le&\exp\left\{-\left(\sum_{t=r+1}^{T}\lambda\gamma^{-(t-r)\rho}\eta+(T-r)((1-\lambda))+r\right)|\Omega_\epsilon|\right\}\\
        =&\exp\left\{-\left((T-r)(\frac{\gamma^{-\rho}+\gamma^{-(T-r)\rho}}{2}\lambda\eta+(1-\lambda))+r\right)|\Omega_\epsilon|\right\}.
    \end{align*}
    In order that $\Pr\left(f(\tilde{x})-f^*>\epsilon\right)\le\delta,$ it suffices that
    \begin{equation*}
        \exp\left\{-\left((T-r)(\frac{\gamma^{-\rho}+\gamma^{-(T-r)\rho}}{2}\lambda\eta+(1-\lambda))+r\right)|\Omega_\epsilon|\right\}\le\delta,
    \end{equation*}
    hence the $(\epsilon,\delta)$-query complexity is upper bounded by 
    \begin{equation*}
        \mathcal{O}\bigg(\max\{\bigl(\frac{\gamma^{-\rho}+\gamma^{-(T-r)\rho}}{2}\lambda\eta+(1-\lambda)\bigr)^{-1}\bigl(\frac{1}{|\Omega_\epsilon|}\ln\frac{1}{\delta}-r\bigr)+r,T\}\bigg).
    \end{equation*}
\end{proof}

\subsection{Sufficient Condition for Acceleration}

Under the assumption that $f$ is dimensionally locally Holder continuous, it is obvious that
\begin{equation*}
    \Omega_\epsilon\subseteq\prod_{i=1}^{n}[x^i_*-(\frac{\epsilon}{L^i_1})^{-\beta^i_1},x^i_*+(\frac{\epsilon}{L^i_1})^{-\beta^i_1}].
\end{equation*}
Denoted by $\tilde{x}_t=(\tilde{x}^1_t,\ldots,\tilde{x}^n_t):=\arg\min_{j=1,\ldots,t}f(x_j)$. The subsequent sufficient condition gives a lower bound of region shrinking rate $\gamma$ and shrinking frequency $\rho,$ such that \emph{RACE-CARS} achieves acceleration over \emph{SRACOS}.

\begin{proposition}\label{hyperparam}{\rm
    For a dimensionally local Holder continuous objective $f.$ Assume that for $\epsilon>0,$ $\Omega_\epsilon$ is $\eta$-shattered by $h_t$ for all $t=r+1,\ldots,T.$ In order that $\Omega_\epsilon$ being $\eta$-shattered by $\tilde{h}_t,$ it is sufficient when the region shrinking rate $\gamma$ and shrinking frequency $\rho$ satisfy:
    \begin{equation*}
        \frac{1}{2}\gamma^{t\rho}\|\Omega\|\ge\biggl(\tilde{x}^1_t-x^1_*+(\frac{\epsilon}{L^1_1})^{-\beta^1_1},\ldots,\tilde{x}^n_t-x^n_*+(\frac{\epsilon}{L^n_1})^{-\beta^n_1}\biggr).
    \end{equation*}
}
\end{proposition}

\end{document}